%% file: main-7401-Yang.tex
\documentclass[11pt,a4paper]{article}
\usepackage{times,latexsym}
\usepackage{url}
\usepackage[T1]{fontenc}

\usepackage[table]{xcolor}
\usepackage{makecell}
\usepackage{amsmath}
\usepackage{amssymb}
\usepackage{hyperref}
\usepackage{adjustbox}
\usepackage{subcaption}
\usepackage{tcolorbox}
\usepackage{multirow}
\usepackage{tabularx}
\usepackage{colortbl}
\usepackage{todonotes}
\usepackage{microtype}
\usepackage{enumitem}
\usepackage{float}
\usepackage{MnSymbol}
\usepackage{booktabs}
\usepackage{multirow}

   \usepackage[acceptedWithA]{tacl2021v1}
\iftaclpubformat 

\else

\fi

\title{Self-Rationalization in the Wild: A Large Scale Out-of-Distribution Evaluation on NLI-related tasks}

\author{
Jing Yang$^{1}$$^{\dagger}$,
Max Glockner$^{2}$,
Anderson Rocha$^{1}$ 
\and
\textbf{Iryna Gurevych}$^{2}$ \vspace{0.25em}\\
  $^1${Artificial Intelligence Lab., \url{Recod.ai}, Institute of Computing, University of Campinas, Brazil} \\
  $^2$UKP Lab,  Department of Computer Science, Technical University of Darmstadt, Germany
  \\
    $^1$\texttt{jing.yang@ic.unicamp.br}, \texttt{anderson.rocha@unicamp.br} \\
  $^2$\texttt{\{max.glockner,iryna.gurevych\}@tu-darmstadt.de} \\
}

\date{}

\begin{document}
\maketitle

\begingroup\renewcommand\thefootnote{$\dagger$}
\footnotetext{Most of work is done during the research stay at UKP.}
\endgroup

\begin{abstract}

    Free-text explanations are expressive and easy to understand, but many datasets lack annotated explanation data, making it challenging to train models for explainable predictions. To address this, we investigate how to use existing explanation datasets for self-rationalization and evaluate models' out-of-distribution (OOD) performance. We fine-tune T5-Large and OLMo-7B models and assess the impact of fine-tuning data quality, the number of fine-tuning samples, and few-shot selection methods. The models are evaluated on 19 diverse OOD datasets across three tasks: natural language inference (NLI), fact-checking, and hallucination detection in abstractive summarization. For the generated explanation evaluation, we conduct a human study on 13 selected models and study its correlation with the Acceptability score (T5-11B) and three other LLM-based reference-free metrics. Human evaluation shows that the Acceptability score correlates most strongly with human judgments, demonstrating its effectiveness in evaluating free-text explanations. Our findings reveal: 1) few annotated examples effectively adapt models for OOD explanation generation; 2) compared to sample selection strategies, fine-tuning data source has a larger impact on OOD performance; and 3) models with higher label prediction accuracy tend to produce better explanations, as reflected by higher Acceptability scores.\footnote{Code available at: \url{https://github.com/UKPLab/tacl2025-ood-eval-self-rationalization}}

\end{abstract}

\input{macros}
\urldef\urlchem\url{https://huggingface.co/t5-base##training-data}
\section{Introduction}

    \begin{figure*}[ht!]
        \centering
        \includegraphics[width=0.99\linewidth]{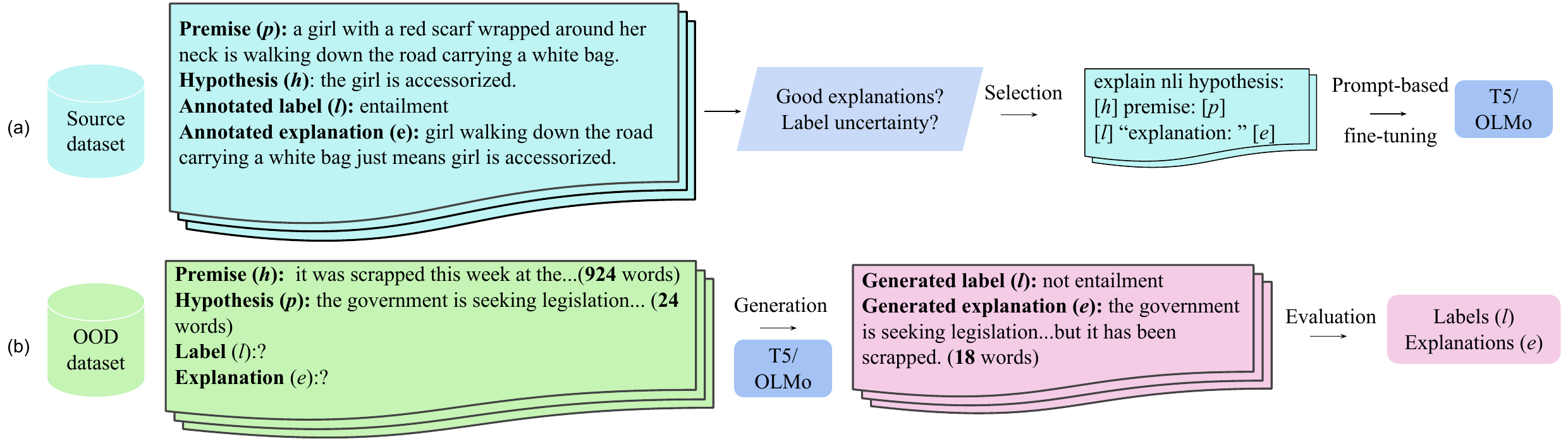}
        \caption{OOD evaluation pipeline of self-rationalization. The pipeline comprises two main parts. The first part (a) relates to \textbf{learning to self-rationalize }with a source dataset (Section \ref{fine-tuning}); it involves sample selection and fine-tuning a generative model. The second part (b) relates to \textbf{OOD generation and evaluation} (Section \ref{ood}); we evaluate the model on three categories of OOD tasks: NLI, fact-checking, and hallucination detection of abstractive summarization.}
        \label{fig:pipeline}
    \end{figure*}

Generating textual explanations has been a major focus in machine learning and NLP~\cite{wei2022chain,kunz2024properties,calderon2024behalf}, as the explanations are expressive and do not require readers to have model-level knowledge to understand. One popular line of work is self-rationalization~\cite{wiegreffe2021measuring,marasovic2022few}, in which a model jointly generates the task label and a free-text explanation for the predicted label. Compared with highlighting words and phrases~\cite{deyoung2020eraser}, free-text explanations can express unstated knowledge and common-sense in easily understandable forms. However, datasets containing annotated free-text explanations are rare due to expensive annotations. 

A few datasets for free-text explanation generation exist~\cite{camburu2018snli,wang2019does,sap2020social,aggarwal2021explanations,chen2022learning}, with e-SNLI~\cite{camburu2018snli} being one of the seminal datasets in the NLI area. Based on SNLI~\cite{bowman2015large}, e-SNLI focuses on reasoning over fine-grained nuances of common-sense knowledge. However, datasets containing longer or more domain-specific text, such as fact-checking on real-world claims, lack annotated explanations~\cite{hanselowski2019richly,saakyan2021covid}. This poses severe challenges for (\emph{i}) training and (\emph{ii}) evaluating self-rationalizing models on these tasks.
No large scale analysis exists to understand how well self-rationalization models can transfer from existing data to unknown~datasets. 

We fill the gap by learning self-rationalization from established sources with annotated explanations and evaluating its generalization performance on 19 out-of-distribution (OOD) datasets over three related tasks (see evaluation setup in Figure~\ref{fig:pipeline}): NLI, fact-checking (FC) and hallucination detection of abstractive summarization (HDAS). NLI focuses on textual entailment within a controlled context, FC extends to real-world claims with retrieved evidence, and HDAS centers around machine-generated text. Our OOD datasets vary in \textit{domains} (e.g., news, Wikipedia, social media, science), and \textit{textual structures} (e.g., synthetic template-based, multiple premises, sentence compositions, long documents), presenting a diverse and challenging OOD~setting (see details of each dataset in Table~\ref{tab:tdata}).

Despite the popularity of LLMs, using them in a large experimental design is prohibitive, as they are computationally expensive to perform inference and evaluation, especially when the input is long. Furthermore, data contamination is a concern when performing evaluations on OOD datasets~\cite{sainz2023nlp}, as the training data of most LLMs are not transparent, such as Llama 2~\cite{touvron2023llama} and GPT-4~\cite{achiam2023gpt}. 
To address this, we selected two open-source models with fully transparent pretraining datasets—T5-Large~\cite{raffel2020exploring} and OLMo-7B~\cite{groeneveld2024olmo}—to study self-rationalization. They also require fewer computational resources than many LLMs, allowing us to perform a large scale study.

We study the impact of fine-tuning data size and quality on OOD performance, focusing on three factors: the fine-tuning dataset, the number of selected samples, and sample selection strategies. To enhance the quality of generated explanations in OOD datasets, we introduce a new approach with an acceptability filtering model~\citep{wiegreffe2022reframing} to select better training samples. 
Our study focuses on fine-tuning instead of in-context learning due to its better OOD generalizability~\cite{mosbach2023shot}. In addition, compared to supervised fine-tuning, in-context learning is subject to additional constraints regarding resources and context window size.

We address the lack of gold reference explanations by studying the effectiveness of the Acceptability score with a human evaluation and comparing it against three LLM-based reference-free metrics. Our results show that Acceptability score correlates highest with humans in all three tasks. 
Our study reveals three findings: 1) fine-tuning on few samples yields comparable OOD performances as fine-tuning on the full dataset; 2) fine-tuning data source has a high impact on OOD performance, while sample selection has a lower impact; 3) higher Acceptability scores are associated with better label prediction performances, providing a new perspective on the task performance vs explainability trade-off.

\section{Related Work}
\paragraph{Free-text explanation generation and evaluation}
Self-rationalization has been a popular approach for generating free-text explanations~\cite{wiegreffe2021measuring,marasovic2022few,ross2022self,veerubhotla2023shot,ramnath2024tailoring}. ~\citet{wiegreffe2021measuring} shows that joint learning of label prediction and explanation generation results in explanations more aligned with predicted labels. ~\citet{marasovic2022few} addressed the scarcity of annotated explanation data by using prompt-based fine-tuning on a few examples, though their evaluation was limited to in-distribution datasets. Few works have studied how such models can generalize to OOD.~\citet{zhou2021investigating} studied how learning with few-shot instances with template-based explanations influences OOD generalization. Their OOD dataset (e-HANS) is limited with constructed templates based on the HANS dataset~\cite{mccoy2019right}.~\citet{ross2022self} studied the effect of self-rationalization on reducing models' reliance on spurious cues in out-of-domain datasets, and they showed that self-rationalization improves models robustness when fine-tuning data size is small. ~\citet{yordano2022few} studied the setup where the target dataset has few annotated free-text explanations but abundant labels. Their approach is limited to target datasets in which free-text explanations exist. In contrast to the above OOD evaluations, we focus on the OOD evaluation of self-rationalization for 19 diverse datasets, and our evaluation does not rely on reference explanations.

Reliable evaluation is crucial for explanation generation. Traditional metrics that measure text overlap with references have shown low correlation with human judgments~\cite{sulem2018bleu}, and reference explanations are not always available. Recent works, like TigerScore~\cite{jiang2023tigerscore}, Auto-J~\cite{li2024generative}, and Themis~\cite{hu2024themis}, use LLMs as evaluators. These metrics rely on detailed instructions specifying evaluation aspects (e.g., relevance, accuracy, coherence) and formatted inputs for the task. The trained metric then generates a rating along with a textual analysis.
To test their suitability for the explanation generated with self-rationalization, in this work, we study their correlations with human judgments.

\paragraph{Few-shot sample selection}
Recent studies show that fine-tuning with smaller, high-quality datasets can outperform larger datasets~\cite{li2024superfiltering,xia2024less}.
~\citet{li2024superfiltering} proposed to use a relatively small language model to evaluate and select a few instances for instruction-tuning on larger models.
To select data to perform well in transfer learning, ~\citet{xia2024less} proposed data selection for instruction-tuning on a target-specific domain. 
They show that training with 5\% of the data outperforms training with the full dataset. The main constraint is that the validation set needs to be from the target domains.
~\citet{chen2024automated} proposed to improve data quality by estimating their model's confidence, and for the low-quality data, they either filter or correct them.
Most methods for sample selection are designed to perform well on in-distribution or known target domains, and the goal is for better classification performance. In contrast, our work focuses on selecting data that should help OOD performance on both label prediction and explanation generation.

\section{Learning to Self-rationalize} \label{fine-tuning}
    
Figure~\ref{fig:pipeline} shows our out-of-distribution (OOD) evaluation pipeline. We first (a) fine-tune a language model on a source dataset to learn self-rationalization. Specifically, we require a fully annotated source dataset $S$, in which each instance contains input $x_s=(h_i, p_i)$ and output $y_s=(l_i, e_i)$, where $h_i, p_i$ represent a hypothesis and premise pair, $l_i$ and $e_i$ represent the annotated label and explanation. We select $m$ representative instances per class from $S$ for fine-tuning by following a sample selection process. Our sample selection method deliberately restrains from using
data from the OOD datasets, preserving them untouched. Finally, we fine-tune a language model to generate a label and explanation. In (b), we evaluate the fine-tuned model performance on OOD datasets (Section \ref{ood}). Given an OOD dataset $O$, with instances $x_o =  (h_j, p_j)$, where $h_j, p_j$ represents a new hypothesis and premise pair, the fine-tuned model generates the label ($\hat{l_j}$) and explanation ($\hat{e_j}$). 

\subsection{Source dataset}

To learn self-rationalization for NLI-related tasks, we select two large source datasets that contain explanations:
    (a) \textbf{e-SNLI}~\cite{camburu2018snli}, derived from the NLI dataset SNLI~\cite{bowman2015large} by adding human annotated explanations. 
    (b) \textbf{e-FEVER}~\cite{stammbach2020fever}, originated from the fact-checking dataset FEVER~\cite{thorne2018fever} with GPT-3 generated synthetic explanations. To improve data quality, we heuristically filter out incorrect explanations from the dataset (see details in Appendix~\ref{append:preprocessing}).
We selected these two datasets as they are representative for our OOD datasets and have abundant explanations.

\subsection{Acceptability-based sample selection} \label{sec:selection}
Inspired by~\citet{schiller2022effect}, we examine how varying the size and quality of fine-tuning data (source dataset) affects OOD performance. Since self-rationalization includes joint label prediction and explanation generation, we propose our method considering both the label and explanation quality:
 
    \paragraph{Data filtering with acceptability score}
    To improve explanation quality, we filter the fine-tuning data using the acceptability model from~\citet{wiegreffe2022reframing}. This model, trained on SNLI data, predicts whether a generated explanation is acceptable based on human judgment. We remove samples with acceptability scores (the predicted probability for the label ``acceptable'') below a 0.3 threshold.

    \paragraph{Data selection}
    
    For data quality estimation in label prediction, we adapt two methods from the literature: (1) \textbf{ambiguous}: Following \citet{swayamdipta2020dataset}, we select samples with high ambiguity, which has been shown to improve OOD generalization. Ambiguity is measured as the distance between an instance's predicted label probability and the mean of all predicted label probabilities using the pre-fine-tuning model (details in Appendix~\ref{append:ambiguous_selection}). (2) \textbf{FastVote-\textit{k}}~\citep{su2022selective}: A graph-based method to select diverse and representative samples. We use the recommended $k=150$.
    
With the combined two steps (data filtering + selection), we denote the sample methods as \textbf{accept-ambiguous} and \textbf{accept-FastVote-\textit{k}}. 

 \subsection{Fine-tuning on source datasets} 

    For fine-tuning T5-Large, we use the standard NLI template~\citep{marasovic2022few}, which has been shown to give the best results for e-SNLI dataset with T5. The encoder and decoder prompts are:    

    \begin{tcolorbox}[colback=gray!5!white,colframe=white!75!black,
                  left=1mm, 
                  right=1mm] 
    \small
    \linespread{1.0}\selectfont
    \textbf{Input}: \textit{explain nli hypothesis:} [hypothesis] \textit{premise:} [premise]
    
    \textbf{Output}: [label] \textit{"explanation: "} [explanation]
    \end{tcolorbox}

    For fine-tuning OLMo-7B, as the model is relatively large, we choose parameter-efficient tuning with LoRA~\cite{hu2022lora} using the following instruction~\cite{zarharan2024tell}. The response is in a JSON format to facilitate extraction of labels and explanations:
    \begin{tcolorbox}[colback=gray!5!white,colframe=white!75!black,
                  left=1mm, 
                  right=1mm] 
    \small
    \linespread{1.0}\selectfont
        \#\#\# Premise: [premise]  Hypothesis: [hypothesis]\\
        \#\#\# Response: \{"relationship": [label], "explanation": [explanation]\}
    \end{tcolorbox}
    
    For the number of shots, we compare 1, 2, 4, 8, 16, 32, 64, and 128 shots. To ensure robustness, we create five subsets from each source dataset, with 5,000 randomly selected samples per subset (with no overlap between subsets). We apply the sample selection methods from Section \ref{sec:selection} to each subset and report the average results (see Appendix \ref{append:ambiguous_selection} for additional fine-tuning details). In total, we fine-tuned 402 T5 models and 302 OLMo~models\footnote{For T5: 2 source datasets ×5 subsets ×8\#shots ×5 sampling methods +2 full-shot models. For OLMo, we discard 1 and 2 shots as our primary results show that models fail to learn with too few examples.}.

 \paragraph{Baselines} We compare the few-shot fine-tuned models with two full-set fine-tuned models on e-SNLI and e-FEVER, respectively. In addition, we include the random sample selection baseline to compare few-shot sample selection methods.

\section{OOD Generation and Evaluation} \label{ood}

This section introduces part (b) of the pipeline in Figure \ref{fig:pipeline}. For all fine-tuned models, we perform inference on all OOD datasets.

\subsection{Out-of-Distribution datasets}
 For a comprehensive evaluation, we collect datasets that resemble the NLI task and divide them into three categories: \textbf{NLI}, Fact-checking (\textbf{FC}), and Hallucination Detection of Abstractive Summarization (\textbf{HDAS}). Table \ref{tab:tdata} lists the OOD datasets used (see Appendix~\ref{append:preprocessing} for dataset details and pre-processing). To ensure no data contamination in our OOD evaluation, we specifically excluded datasets used for supervised fine-tuning of~T5~\citep{raffel2020exploring}. OLMo model was pre-trained on Dolma~\cite{soldaini2024dolma} corpus, which contains data from diverse sources but is not fine-tuned with curated NLI datasets. 

\begin{table*}[!ht]
\centering
\begin{adjustbox}{width=1.0\linewidth}
\begin{tabular}{llrrlrrr}
\toprule
\multicolumn{1}{l}{}    & \textbf{OOD dataset} & \multicolumn{1}{r}{\textbf{Size}} & \multicolumn{1}{r}{\textbf{\#L.}} & \textbf{Domain} & \multicolumn{1}{r}{\textbf{\begin{tabular}[r]{@{}r@{}}\#words \\ (Hyp.)\end{tabular}}} & \multicolumn{1}{r}{\textbf{\begin{tabular}[r]{@{}r@{}}\#words \\ (Pre.)\end{tabular}}} & \makecell{\textbf{IAA}}\\ 
\midrule
\rowcolor[HTML]{FFF2CC}  & SICK \cite{marelli2014sick}   & 4,906    & 3  & {news, image captions} & 10   & 10    & 0.84$^O$   \\
\rowcolor[HTML]{FFF2CC}   & {AddOneRTE \cite{pavlick2016most}}              & 387      & 2   & {news, image captions, forums, literature}  & 13     & 12     &  0.77$^O$ \\
\rowcolor[HTML]{FFF2CC}  & {JOCI~\cite{zhang2017ordinal}}   & 39,092     & 3  &{image captions,  commonsense stories}  & 6  & 14   & 0.54$^C$  \\
\rowcolor[HTML]{FFF2CC}    & {MPE \cite{lai2017natural}}                     & 1,000     & 3 & image captions & 4    & 48   & 0.70$^O$   \\
\rowcolor[HTML]{FFF2CC}   & {DNC \cite{poliak2018collecting} }                    & 60,036   & 2  & {events, named entities, puns, sentiments}  & 5    & 19   & -   \\
\rowcolor[HTML]{FFF2CC}   & {HANS \cite{mccoy2019right}}  & 30,000     & 2  & {template-based (synthetic)} & 6       & 9  &-  \\
\rowcolor[HTML]{FFF2CC}   & {WNLI \cite{wang2019glue} }   & 71    & 2   & fiction books & 7  & 21  &- \\
\rowcolor[HTML]{FFF2CC}   & {Glue Diagnostics \cite{wang2019glue}}    & 1,104   & 3 & {news, Reddit, Wikipedia, academic papers}   & 16   & 16  &0.73$^F$   \\
\rowcolor[HTML]{FFF2CC} 
\multirow{-9}{*}{\rotatebox{90}{NLI}}    & {ConjNLI \cite{saha2020conjnli}}  & 623   & 3   & Wikipedia  & 13    & 13   & 0.83$^C$ \\ 
\midrule
\rowcolor[HTML]{F4CCCC}   & {Snopes Stance \cite{hanselowski2019richly}}     & 1651   & 3  & {Snopes (fact-checking platform)} & 16   & 126  & 0.70$^C$ \\
\rowcolor[HTML]{F4CCCC}   & {SciFact \cite{wadden2020fact}}   & 300   & 3   & {biomedicine,  scientific articles} & 13   & 247  &0.75$^C$ \\
\rowcolor[HTML]{F4CCCC}   & {Climate-FEVER \cite{diggelmann2020climate}}           & 1,381      & 3  & {climate change, Google searches} & 20   & 136   &0.33$^K$ \\
\rowcolor[HTML]{F4CCCC}   & {VitaminC \cite{schuster2021get}}    & 55,197  & 3  & {Wikipedia, COVID-19}  & 13  & 28 & 0.71$^F$ \\
\rowcolor[HTML]{F4CCCC}  & {COVID-FACT \cite{saakyan2021covid}}   & 4,086  & 2 & {Reddit, COVID-19} & 12  & 73 & 0.50$^C$\\
\rowcolor[HTML]{F4CCCC} 
\multirow{-6}{*}{\rotatebox{90}{FC}} & {FM2 \cite{eisenschlos2021fool}}    & 1,380   & 2  & {Wikipedia} & 14   & 32 & -\\ 
\midrule
\rowcolor[HTML]{C9DAF8}  & {FactCC \cite{kryscinski2020evaluating}}  & 503 & 2 & {news (CNN/DailyMail), rule-based} & 14  & 644  &0.75$^C$ \\
\rowcolor[HTML]{C9DAF8}  & {QAGs CNNDM \cite{wang2020asking}}              & 714 & 2  & {news (CNN/DailyMail), BART-based}  & 16  & 318 & 0.51$^K$  \\
\rowcolor[HTML]{C9DAF8}  & {QAGs XSUM \cite{wang2020asking}}                & 239   & 2  & {news (XSUM), BART-based} & 18  & 351 & 0.34$^K$  \\
\rowcolor[HTML]{C9DAF8} 
\multirow{-4}{*}{\rotatebox{90}{HDAS}}   & {XSUM Hallucination \cite{maynez2020faithfulness}}  & 1,869 & 2 & {news (XSUM), 7 different models} & 19 & 361 & 0.92$^O$  \\ 
\bottomrule
\end{tabular}
\end{adjustbox}
\caption{OOD datasets categories and details. NLI: yellow, FC: pink, and HDAS: blue. Hyp.: hypothesis, Pre.: premise, \#words: number of words in average, IAA: inter-annotator agreement (numbers are from the original papers). L.: labels, $C$: Cohen’s kappa, $F$: Fleiss’s kappa, $K$: Krippendorff’s alpha, $O$: other metrics, -: unspecified. The sizes are reported on test/dev split; if the split is not provided, we report and evaluate on the entire dataset.}
\label{tab:tdata}
\end{table*}

    \paragraph{NLI}
    NLI datasets access models' ability to infer relationships between sentences, with challenges ranging from compositional meaning \cite{marelli2014sick}, adjective-noun composition \cite{pavlick2016most}, common-sense inference \cite{zhang2017ordinal}, to multiple premise entailment \cite{lai2017natural}. DNC~\cite{poliak2018collecting} expands the challenge by incorporating diverse semantic phenomena into the NLI format. HANS~\cite{mccoy2019right} and WNLI~\cite{wang2019glue} are two adversarial datasets designed to reveal models' underlying heuristic biases. Glue Diagnostics~\cite{wang2019glue} and ConjNLI~\cite{saha2020conjnli} further diversify the NLI task, testing models against a wide array of linguistic challenges and over conjunctive~sentences. 
    
    \paragraph{FC}  
    FC datasets aim to evaluate the veracity of claims against evidence from various sources and topics, including fact-checking platforms~\cite{hanselowski2019richly}, scientific articles~\cite{wadden2020fact}, Wikipedia~\cite{schuster2021get,eisenschlos2021fool}, and information related to climate change and COVID-19~\cite{diggelmann2020climate,saakyan2021covid}. These datasets require models to evaluate the truthfulness of claims in real-world scenarios from very different domains. To avoid error propagation, we use gold evidence for all FC datasets.

    \paragraph{HDAS}
    HDAS datasets encompass a variety of model-generated summaries, reflecting the evolving landscape of automatic text generation and its implications for information integrity. FactCC~\cite{kryscinski2020evaluating} challenges models to identify inaccuracies in summaries generated through five rule-based transformations. QAGS CNN and QAGS XSUM~\cite{wang2020asking}, derived from CNN/DailyMail and XSUM datasets, consist of summaries generated by the BART model~\cite{lewis2020bart}. XSUM Hallucination~\cite{maynez2020faithfulness} contains factuality annotated summaries generated by seven models.

In comparison, the three tasks vary in objective, domain, and text length. NLI targets logical relationships between sentences, requiring models to handle linguistic subtleties and logic-based reasoning in a controlled textual context. FC focuses on real-world applicability, requiring external information and complex reasoning between sentences and documents. HDAS addresses the problems of automatic document summarization. Regarding text length, FC datasets typically have longer premises than NLI, with HDAS having the longest. Together, these datasets present a challenging NLI-related OOD scenario.

\subsection{Inference on OOD datasets } 

During OOD inference, fine-tuned models may not generate a label and explanation following the output template. To address this, for T5 models, we take the first token to represent the predicted label. 
For datasets that only include two classes (``entailment'' and ``non-entailment''), we merge the ``contradiction'' and ``neutral'' labels into the ``non-entailment'' label (see more details on label extraction in Appendix~\ref{append:implementation_details}).
We detect explanations by searching for the pattern ``explanation: '' and, if absent, treat all text after the first word as the explanation.
For OLMo models, as we instruction-tuned the model to generate a JSON-formatted output, we extract the labels and explanations by finding their keys and if not found, we set both to be none.
        
\section{Results and Analysis}
This section begins with OOD label prediction results. We then evaluate explanations through human judgments and analyze their correlation with reference-free metrics. Next, we report explanation evaluation results across all datasets using the most correlated metric. Finally, we present the overall performance of each OOD dataset on the best-performing models.

\subsection{OOD Performance on Label Prediction}

\begin{figure*}[t]
    \centering
    \includegraphics[width=\linewidth]{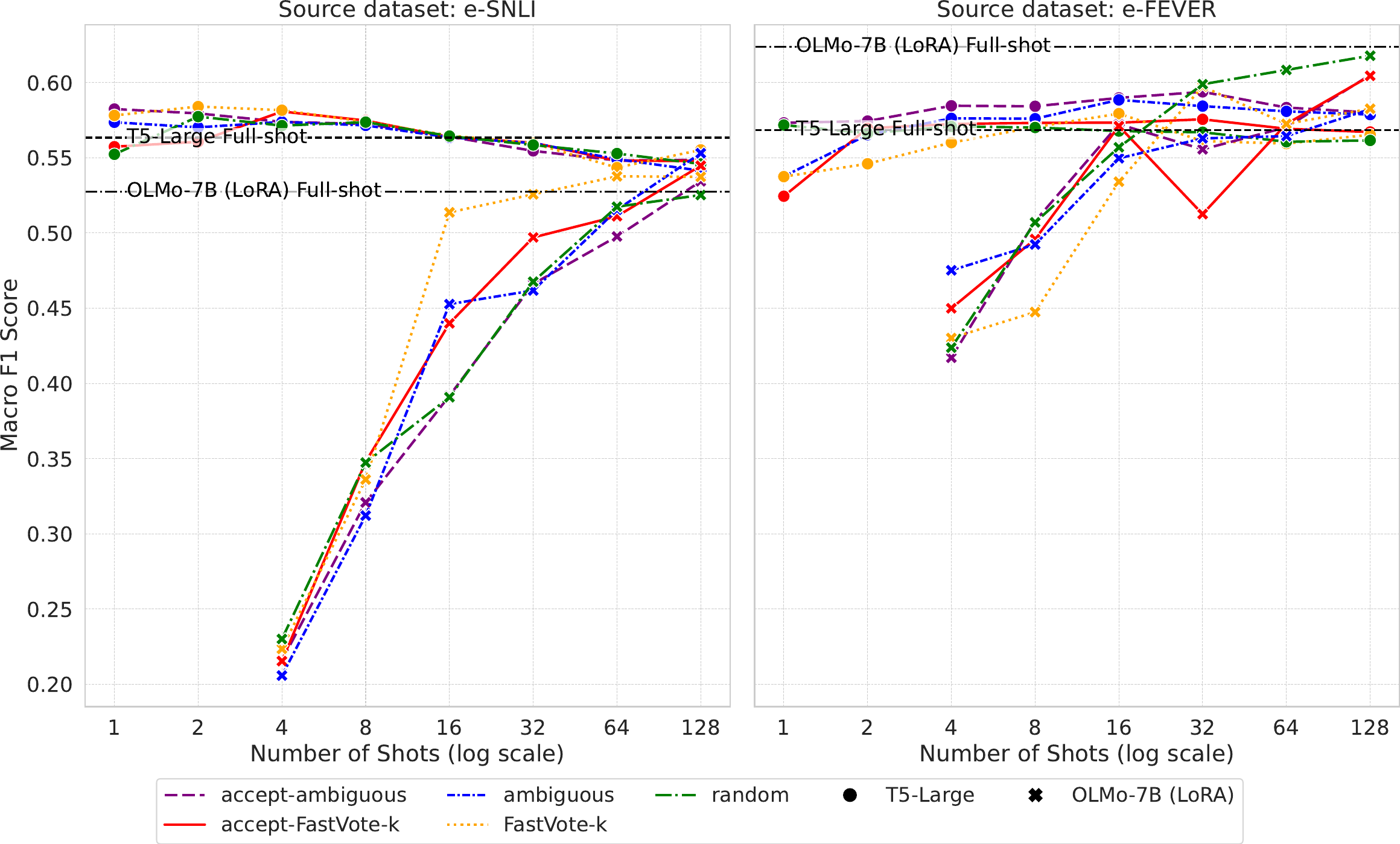}
    \caption{Average Macro F1 score across different number of shots and sample selection~methods. Each point is the average of all 19 OOD datasets, and 5 models from the 5 subsets.}
    \label{fig:acc_all}
\end{figure*}

We compare the OOD label prediction performance of fine-tuned T5-Large and OLMo-7B models on two source datasets, considering various sample selection methods and number of shots, as shown in Figure \ref{fig:acc_all}. Label prediction performance is measured using the Macro F1 score. 

\paragraph{T5 vs. OLMo:} As shown in Figure \ref{fig:acc_all}, T5 and OLMo models exhibit distinct trends in label prediction performance as the number of shots increases. OLMo starts with low performance, improving almost monotonically with more shots. T5, however, shows less variation, starting with slightly higher performance and then reaching levels similar to full-shot models. This difference may be because of T5's pre-training on NLI datasets (MNLI, QNLI, RTE, CB), allowing it to handle NLI tasks effectively without much benefit from additional fine-tuning (see detailed discussion in Section~\ref{sec:source-f1}). This is further indicted by the results: T5 full-shot fine-tuning with both source datasets have similar F1 scores, and neither yields better results than their best few-shot counterparts. 

\paragraph{e-SNLI vs. e-FEVER:} Overall, e-FEVER models achieve better average OOD F1 scores than e-SNLI, and the OLMo model fine-tuned on e-FEVER full-shot has the highest OOD F1 score. For e-SNLI, T5 and OLMo models reach similar performances at 128 shots, but the trends are the opposite. For e-FEVER, the performance of T5 models tends to stabilize after only 2-shots, while the performance of OLMo models continues to increase and eventually surpass T5 models.

\paragraph{Sample Selection}

 As depicted in Figure~\ref{fig:acc_all}, no sample selection method consistently outperforms others in label prediction. For T5 models, selection methods perform similarly, especially in e-SNLI; although ``accept-ambiguous'' method is slightly better in e-FEVER. For OLMo models, ``FastVote-\textit{k}'' excels in e-SNLI, while ``random selection'' achieves slightly higher scores than others in e-FEVER (after 32 shots), nearly matching full-shot performance. Surprisingly, ``FastVote-\textit{k}'' and ``ambiguous'' do not surpass the random baseline, possibly due to outliers and training instability when using small numbers of samples~\cite{karamcheti2021mind,su2022selective}.

\subsection{OOD Explanation Quality Evaluation} \label{ref:human_eval}

We evaluate the generated explanations using both human evaluation and reference-free automatic metrics, and analyze the correlation between them.

\subsubsection{Human evaluation setup}

\begin{table}[t]
\centering
\begin{adjustbox}{width=1.0\linewidth}
\begin{tabular}{lllrll}
\toprule
Acronym & Source & Model & \#Shots & Selection \\
\midrule
T$^{Fev}_{64,AFk}$ & e-FEVER & T5 & 64 & accept-FastVote-\textit{k}\\
T$^{Fev}_{128,R}$ & e-FEVER & T5 & 128 & random \\
T$^{Fev}_{128,Fk}$& e-FEVER & T5 & 128 & FastVote-\textit{k} \\
T$^{Fev}_{128,AFk}$ & e-FEVER & T5 & 128 & accept-FastVote-\textit{k} \\
T$^{Fev}_{Full}$ & e-FEVER & T5 & Full & - \\
\midrule
T$^{Sn}_{64,Fk}$ & e-SNLI & T5 & 64 & FastVote-\textit{k} \\
T$^{Sn}_{64,AFk}$ & e-SNLI & T5 & 64 & accept-FastVote-\textit{k} \\
T$^{Sn}_{Full}$ & e-SNLI & T5 & Full & - \\
\midrule
O$^{Fev}_{16,AFk}$ & e-FEVER & OLMo & 16 & accept-FastVote-\textit{k} \\
O$^{Fev}_{128,AFk}$ & e-FEVER & OLMo & 128 & accept-FastVote-\textit{k} \\
O$^{Fev}_{Full}$  & e-FEVER & OLMo & Full & -\\
\midrule
O$^{Sn}_{128,AFk}$ & e-SNLI & OLMo & 128 & accept-FastVote-\textit{k} \\ 
O$^{Sn}_{Full}$ & e-SNLI & OLMo & Full & -\\
\bottomrule
\end{tabular} 
\end{adjustbox}
\caption{Selected models for human evaluation for the models \textbf{T}5 and \textbf{O}LMo. The left most column shows the acronym of the models, which will be used throughout the rest of the paper.}
\label{tab:models}
\end{table}

Conducting a human study is challenging due to the extensive number of models and OOD datasets. Thus, we select three OOD datasets (SICK, VitaminC, XSUM Hallucination) representing NLI, FC, and HDAS, respectively. 
To study the impact of fine-tuning factors on OOD explanations, we select models that demonstrated high and comparable F1 scores averaged across the three OOD datasets (see Figure~\ref{fig:f1_3} in Appendix~\ref{append:comple_results} with the selected models highlighted). Table~\ref{tab:models} lists the 13 selected mode details, with first column provides models' acronyms for across reference later (examples of generated explanations by the selected models can be found in Table~\ref{tab:exam_exps_sick},~\ref{tab:exam_exps_vc} and~\ref{tab:exam_exps_sxum} in Appendix~\ref{append:generated_exps}). We use the Prolific platform for recruiting workers, and the open-source POTATO annotation tool~\cite{pei2022potato} for the evaluation interface.

For instance selection, following~\citet{marasovic2022few}, we shuffle each dataset and select the first 15 correctly predicted instances per class and model. This results in 1,560 instances, including those with identical hypothesis-premise pairs but different model-generated explanations. Each instance is evaluated by three different workers, and each worker evaluates 10 instances, requiring in total 468 crowd-workers. Evaluators are shown the hypothesis-premise pair, the gold label, and the generated explanation. Their task is to answer two questions (see the evaluation page in Figure \ref{fig:example_eval} of Appendix~\ref{append:human_evaluation}).
        \begin{myitemize}
            \item Given the Hypothesis and Premise, does the Explanation justify the given Relationship (Single-selection)? Options: \textit{Yes}, \textit{Weakly Yes}, \textit{Weakly No} and \textit{No}.
            \item What are the shortcomings of the Explanation (Multi-selection)? Options: \textit{Does not make sense}, \textit{Insufficient justification},  \textit{Irrelevant to the task}, \textit{Too trivial (only repeating one of the sentences)}, \textit{Contains hallucinated content (not present the premise)} and \textit{None (only if the previous answer is Yes)}. 
        \end{myitemize}

We calculate the average score for each instance from 3 evaluators by assigning the weight to the selected answers as follows~\cite{marasovic2022few,yordano2022few}: Yes: 1, Weakly Yes: 2/3, Weakly No: 1/3 and No: 0. 

\subsubsection{Evaluation with reference-free metrics}

 Originally designed for filtering GPT-3 generated NLI explanations, we propose to use the \textbf{Acceptability score}\footnote{In this paper, when mentioning the acceptability filter (T5-Large), we start with lowercase ``a'', and the Acceptability metric (T5-11B) capital ``A''.}~\cite{wiegreffe2022reframing} as a reference-free metric for explanation evaluation. We choose the largest size of the model variance: T5-11B, which assigns a score between 0 and 1. We compare the metric against state-of-the-art NLG reference-free evaluation metrics:

    \begin{myitemize}
        \item \textbf{Auto-J}~\cite{li2024generative}: trained with LLaMA-2-13B-chat model to evaluate LLM-generated responses. The metric generates an explanation for its judgment and a final integer rating from 1 to~10.

        \item \textbf{TigerScore}~\cite{jiang2023tigerscore}: trained with LLaMA-2 on MetricInstruct dataset. We choose the larger size of the metric: TIGERScore-13B. It generates a breakdown error analysis and a final error score from 0 to -infinity (the smaller, the better).
        
        \item \textbf{Themis}~\cite{hu2024themis}: trained with Llama-3-8B based on their constructed dataset NLG-Eval. It offers flexible aspect-based evaluations across different tasks. We tested three aspects—relevance, coherence, and consistency—and selected relevance due to its highest correlation with human judgments. The metric outputs an evaluation analysis and provides a scale rating from~1~to~5.
        \end{myitemize}

For all reference-free metrics, we calculate the scores for all samples in the datasets, given ground truth inputs (hypothesis, premise, and gold label). Appendix~\ref{append:eval_template} presents the instructions of the evaluation models.

    \subsubsection{Correlation between human evaluation and automatic evaluation metrics}

  \begin{table}
    \centering
    \footnotesize
    \begin{tabular}{lrrrr}
    \toprule
    Dataset & Auto-J  & TigerScore  & Themis  & Accept. \\
    \midrule
    SICK & -0.011 & -0.220 & 0.400 & \textbf{0.466} \\
    VitaminC & 0.163 & -0.263 & 0.394 & \textbf{0.469} \\
    XSUM H. & 0.223 & -0.216 & 0.326 & \textbf{0.475} \\
    \midrule
    All & 0.123 & -0.219 & 0.387 & \textbf{0.484} \\
    \bottomrule
    \end{tabular} 
    \caption{Spearman's correlation between human scores and automatic scores in different OOD datasets. All correlation coefficients are significant with $\rho$ < 0.001, except for Auto-J on SICK.}
    \label{tab:correlation_datasets}
    \end{table}
    
    Table \ref{tab:correlation_datasets} shows the Spearman's correlation\footnote{Unlike Spearman, Pearson correlation assumes variables to be continuous and from a normal distribution.} between human and reference-free metrics for the three OOD datasets.     
    The Acceptability score (T5-11B) has the highest correlation with human evaluation for all datasets, followed by Themis, and Auto-J has the lowest. The highest correlations in all three datasets demonstrate the usability of the Acceptability score as a reference-free metric for the explanation evaluation of NLI-related tasks.

    \subsubsection{Evaluation results on selected models and instances}

    The average scores of human evaluations in the three OOD datasets are shown in Table~\ref{tab:scores_datasets} in Appendix~\ref{append:comple_results}. The scores show that SICK has the highest explanation scores, with VitaminC slightly lower than SICK's, and XSUM Hallucination the lowest, agreed by humans and two automatic metrics. This may be due to the extremely long premise/document in the XSUM dataset, making it difficult for the model to generate good explanations. For shortcomings of explanations, see the detailed results in Figure~\ref{fig:answer-reasons} in Appendix \ref{append:comple_results}).  

    Table~\ref{tab:n_shots} shows the evaluation results on the 13 selected models. We include Acceptability and Themis scores as they have moderate correlations with humans. In addition, we show the average Acceptability score on all 19 datasets for overall results. In comparison with Table~\ref{tab:correlation_datasets}, the Acceptability scores in general agree with Humans when the differences between models are large, although the highest scores are not always agreed by all metrics. We discuss the evaluation results regarding each factor in the following. 
    
\begin{table}[!ht]
\centering
\begin{adjustbox}{width=1.0\linewidth}
\begin{tabular}{lccc|c}
\toprule
\textbf{Model} & \textbf{Human} & \textbf{Themis} & \textbf{Accept. (3)} & \textbf{Accept. (19)} \\
\midrule
T$^{Fev}_{64,AFk}$ & 0.631 & 2.058 & 0.317 & 0.250 \\
T$^{Fev}_{128,R}$ & 0.623 & 1.983 & 0.276 & 0.206 \\
T$^{Fev}_{128,Fk}$ & 0.589 & 1.867 & 0.216 & 0.201 \\
T$^{Fev}_{128,AFk}$ & 0.611 & \textbf{2.092} & \textbf{0.328} & \textbf{0.256} \\
T$^{Fev}_{Full}$ & \textbf{0.653} & 1.958 & 0.309 & 0.191 \\
\midrule
T$^{Sn}_{64,Fk}$ & 0.621 & 2.133 & 0.369 & 0.259 \\
T$^{Sn}_{64,AFk}$ & \textbf{0.679} & \textbf{2.367} & 0.418 & 0.281 \\
T$^{Sn}_{Full}$ & 0.678 & 2.050 & \textbf{0.519} & \textbf{0.343} \\
\midrule
O$^{Fev}_{16,AFk}$ & 0.631 & \textbf{2.417} & \textbf{0.423} & 0.305 \\
O$^{Fev}_{128,AFk}$ & 0.639 & 2.250 & 0.384 & \textbf{0.307} \\
O$^{Fev}_{Full}$ & \textbf{0.656} & 1.917 & 0.311 & 0.219 \\
\midrule
O$^{Sn}_{128,AFk}$ & \textbf{0.643} & \textbf{2.300} & \textbf{0.491} & \textbf{0.303} \\
O$^{Sn}_{Full}$ & 0.408 & 1.208 & 0.194 & 0.111 \\
\bottomrule
\end{tabular}
\end{adjustbox}
\caption{Evaluation results on OOD datasets of the 13 selected models. 3 means on the three selected datasets, 19 means all datasets. Models are grouped by base models and source datasets.}
\label{tab:n_shots}
\end{table}

\paragraph{T5 vs OLMo} Table~\ref{tab:n_shots} shows that the difference between the two base models is most pronounced with e-SNLI full-shot. T5 fine-tuned on full shot e-SNLI (T$^{Sn}_{Full}$) provides the best explanations (besides T$^{Sn}_{64, AFk}$), whereas OLMo on full-shot e-SNLI (O$^{Sn}_{Full}$) generates the worst explanations. This may be due to catastrophic forgetting~\cite{luo2023empirical} in the OLMo model when fine-tuned on too many e-SNLI samples, causing the model's reduced OOD generalization ability. 

\paragraph{e-SNLI vs e-FEVER}  Most e-SNLI models achieved higher scores than e-FEVER in explanation quality (under the same model type and number of shots), except for OLMO full-shot. This could be attributed to the higher quality of explanations in the e-SNLI source dataset, while e-FEVER explanations are generated by GPT-3 (see more detailed comparison in Section~\ref{sec:source-accept}).

\paragraph{Few vs Full} Overall, few-shot models achieved similar human scores to their full-shot counterparts, except for the OLMo full-shot e-SNLI model. Although full-shot models showed slightly higher human scores, reference-free metrics favored the explanations generated by few-shot models, particularly for e-FEVER models. 

\paragraph{Sample Selection} As shown in Table~\ref{tab:n_shots}, using the acceptability filter (``accept-FastVote-\textit{k}'') improves explanation quality compared with the same sample selection without the filter (``FastVote-\textit{k}''); however, T$^{Fev}_{128,AFk}$ is not better than random selection (T$^{Fev}_{128,R}$) according to humans.
Nevertheless, based on the scores from the two reference-free metrics, using the acceptability filter improves generated explanation quality (see more detailed discussion in Section~\ref{sec:source-accept}). 

\subsection{Self-Rationalization in the Wild: Overall OOD Performance}

A good self-rationalization model should perform well both on label prediction and explanation generation. Thus, we first evaluate the generated explanations from a large number of models using the Acceptability score (for all instances, we use the gold labels for calculating the Acceptability score). Due to computational constraints, we limit the number of shots to 4, 16, 64, 128, and full, with data selected from the first subset (the Acceptability scores across different number of shots and sample selections can be found in Figure~\ref{fig:accept_all} of Appendix~\ref{append:comple_results}). We then show models' overall performance considering both the F1 and Acceptability score. Finally, we select the best-performing models to demonstrate overall performance on the 19 OOD~datasets.

\subsubsection{Relationship between label prediction performance and explanation quality}

\begin{figure*}[t]
    \centering
\includegraphics[width=\linewidth]{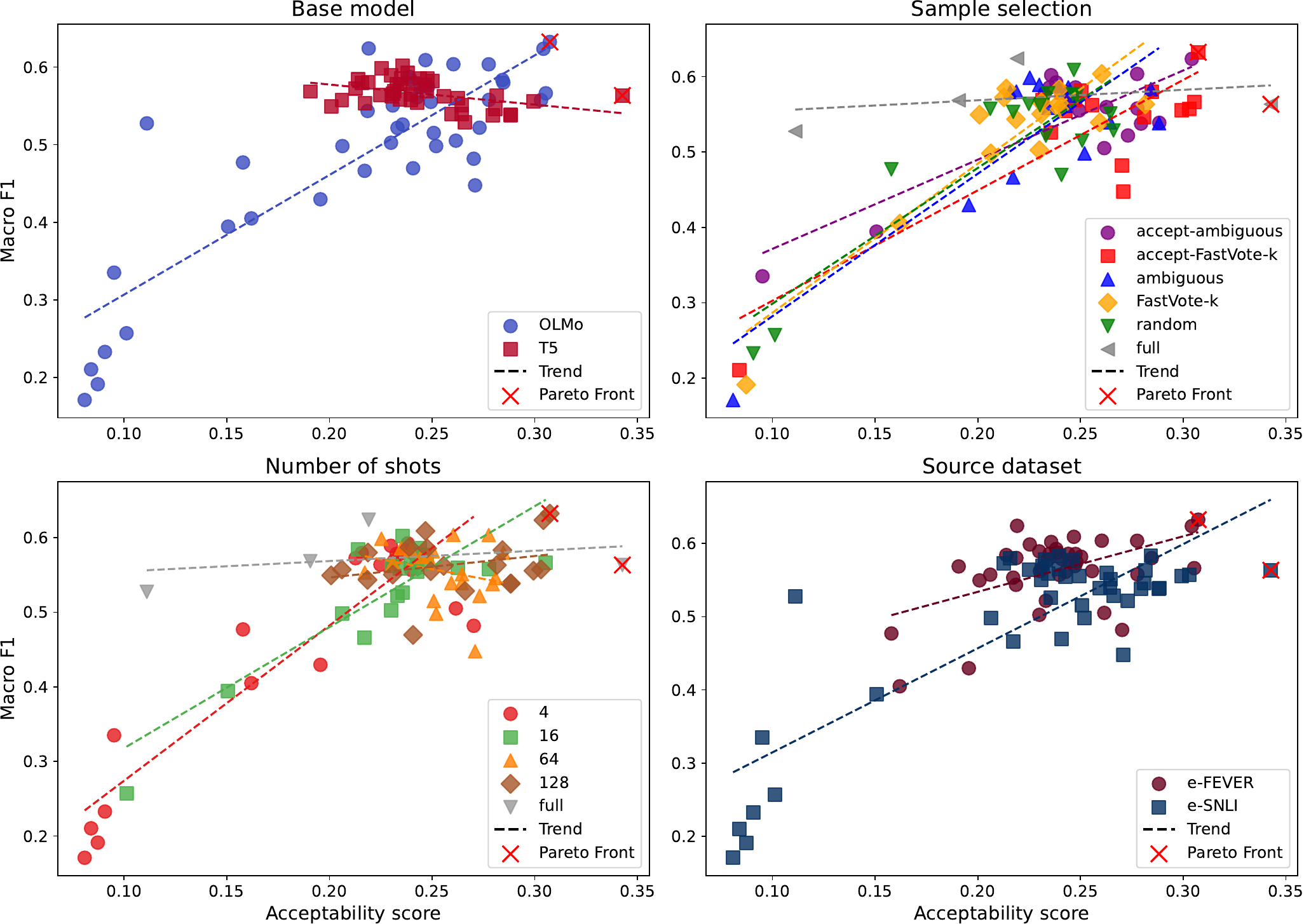}
    \caption{Distribution of models under different fine-tuning factors, with the x-axis showing the Acceptability score, and the y-axis the macro F1 score (scores are averaged over all datasets). The dashed lines are the estimated linear trends of the Acceptability score and macro F1 score.}
    \label{fig:accept_f1}
\end{figure*}

Figure \ref{fig:accept_f1} shows the distribution of models under different fine-tuning factors, with the x-axis showing the Acceptability score and the y-axis the macro F1 score (scores are averaged over all datasets). We select the best models based on the Pareto fronts\footnote{For each point if no other point is strictly higher in both scores, the point is part of the Pareto front~\cite{ben1980characterization}.}. 

As depicted in Figure~\ref{fig:accept_f1}, higher Acceptability scores are usually associated with better F1 scores. Regarding each factor, we see that 1) OLMo models' OOD performances are less stable than T5 models' but achieve better results with higher numbers of shots; 2) Sample selection methods with the acceptability filter have higher Acceptability scores; 3) Comparing the source datasets, fine-tuning on e-SNLI in general achieve higher Acceptability scores while on e-FEVER yield better F1 scores (see more discussions on the impact of each factor in Section~\ref{sec:discuss}).

Regarding the best-performing models that consider both labels and explanations, two models are selected based on the Pareto front: O$^{Fev}_{128,AFk}$ (OLMo, 128 shots, accept-Fastvote-\textit{k}, e-FEVER) and T$^{Sn}_{Full}$ (T5, full-shot, e-SNLI). The first achieves the highest F1 score, while the second has the best Acceptability score, with both models performing competitively on the other metric.

\subsubsection{Performance on the 19 OOD Datasets}

 Table \ref{tab:all_ood} shows the F1 score and Acceptability score on the best models across each OOD dataset (state-of-the-art results on each dataset can be found in Table~\ref{tab:acc_comparison} of Appendix \ref{append:comple_results}). As a comparison, we also include two other models with the same configurations as the best models but trained on a different source dataset: T$^{Fev}_{Full}$ and O$^{Sn}_{128,AFk}$.

Table~\ref{tab:all_ood} shows that the O$^{Fev}_{128,AFk}$ model achieves the highest F1 score on most OOD datasets, though its Acceptability score is slightly lower than that of the T$^{Sn}_{Full}$ model. When comparing e-SNLI and e-FEVER fine-tuned models, e-FEVER models generally perform better in F1 scores on FC and HDAS datasets, with O$^{Fev}_{128,AFk}$ scoring about 10 percentile higher on average for FC (slightly less) and HDAS (slightly more). In terms of explanation generation, OLMo-based models exhibit better performance. Even on e-FEVER, OLMo achieves competitive scores across most OOD datasets, whereas the T5 model fine-tuned on e-FEVER (T$^{Fev}_{Full}$) produces the worst explanations, except for the HDAS task (this might also be due to the number of shots difference, as fine-tuned on more number of shots with e-FEVER do not always lead to better explanations). Finally, the Acceptability scores show a decreasing trend from NLI to HDAS tasks, consistent with previous human evaluation results (see Table~\ref{tab:scores_datasets} in Appendix~\ref{append:comple_results}), where datasets with longer premises generally resulted in lower Acceptability~scores.

\begin{table*}[ht!]
    \centering
    \footnotesize
    \begin{tabular}{lrrrr|rrrr}
        \toprule
        & \multicolumn{4}{c|}{\textbf{Macro F1 score}} & \multicolumn{4}{c}{\textbf{Acceptability score}} \\ 
        \midrule
        \textbf{Dataset} & \textbf{T$^{Sn}_{Full}$} & T$^{Fev}_{Full}$ & O$^{Sn}_{128,AFk}$ & \textbf{O$^{Fev}_{128,AFk}$} & \textbf{T$^{Sn}_{Full}$} & T$^{Fev}_{Full}$ & O$^{Sn}_{128,AFk}$ & \textbf{O$^{Fev}_{128,AFk}$} \\ 
        \midrule
        \rowcolor[HTML]{FFF2CC} SICK & 58.5 & \textbf{78.8} & 55.4 & \underline{65.1} & \textbf{53.0} & 18.5 & \underline{47.5} & 40.2 \\
        \rowcolor[HTML]{FFF2CC} AddOneRTE & \underline{72.3} & \textbf{75.6} & 65.0 & 72.0 & \underline{44.5} & 9.3 & \textbf{44.9} & 39.4 \\
        \rowcolor[HTML]{FFF2CC} JOCI & \underline{52.5} & 41.8 & 49.2 & \textbf{53.7} & \textbf{51.9} & 12.4 & \underline{43.6} & 41.6 \\
        \rowcolor[HTML]{FFF2CC} MPE & \textbf{68.7} & 37.7 & \underline{62.4} & 60.7 & \textbf{49.8} & 6.4 & \underline{45.8} & 39.2 \\
        \rowcolor[HTML]{FFF2CC} DNC & \underline{60.1} & \textbf{66.9} & 53.4 & 58.5 & \textbf{35.1} & 10.0 & 25.8 & \underline{32.8} \\
        \rowcolor[HTML]{FFF2CC} HANS & \underline{58.2} & 43.3 & 51.7 & \textbf{65.9} & \textbf{38.6} & 27.6 & 24.0 & \underline{27.8} \\
        \rowcolor[HTML]{FFF2CC} WNLI & 35.0 & 32.4 & \underline{42.1} & \textbf{55.1} & \underline{29.9} & 22.7 & \textbf{31.7} & 28.0 \\
        \rowcolor[HTML]{FFF2CC} Glue Diagnostics & 57.9 & \underline{59.3} & 57.7 & \textbf{61.3} & \textbf{47.9} & 29.0 & \underline{42.7} & 41.9 \\
        \rowcolor[HTML]{FFF2CC} Conj & \underline{62.6} & \textbf{65.4} & 58.1 & 56.9 & \textbf{48.7} & 30.4 & \underline{41.4} & 38.7 \\
        \rowcolor[HTML]{F4CCCC} Snopes Stance & 36.8 & 44.1 & \underline{45.7} & \textbf{58.4} & \textbf{20.1} & 9.9 & 18.1 & \underline{20.1} \\
        \rowcolor[HTML]{F4CCCC} SciFACT & 60.7 & \underline{62.5} & 56.2 & \textbf{70.0} & \underline{25.7} & 17.6 & 22.5 & \textbf{25.8} \\
        \rowcolor[HTML]{F4CCCC} Climate FEVER & 46.9 & \underline{47.5} & 42.4 & \textbf{51.3} & \textbf{20.9} & 12.8 & 18.4 & \underline{20.8} \\
        \rowcolor[HTML]{F4CCCC} VitaminC & 55.8 & \textbf{58.8} & 55.3 & \underline{56.5} & \textbf{40.3} & 29.8 & \underline{39.2} & 37.2 \\
        \rowcolor[HTML]{F4CCCC} COVID-Fact & 63.3 & \underline{65.9} & 55.3 & \textbf{69.8} & \textbf{28.1} & 12.2 & 19.8 & \underline{23.5} \\
        \rowcolor[HTML]{F4CCCC} FM2 & 70.2 & 71.7 & \underline{76.0} & \textbf{79.3} & \underline{38.4} & 24.1 & \textbf{39.0} & 38.1 \\
        \rowcolor[HTML]{C9DAF8} FactCC & 56.4 & \underline{59.6} & 56.0 & \textbf{65.2} & 16.8 & \textbf{27.6} & 19.1 & \underline{24.6} \\
        \rowcolor[HTML]{C9DAF8} QAGS CNN & 51.8 & 59.3 & \underline{60.0} & \textbf{72.5} & 20.2 & \textbf{26.4} & 19.0 & \underline{25.8} \\
        \rowcolor[HTML]{C9DAF8} QAGS XSUM & 55.0 & 59.3 & \underline{61.4} & \textbf{72.6} & \textbf{24.0} & 15.9 & 19.0 & \underline{23.0} \\
        \rowcolor[HTML]{C9DAF8} XSUM H. & 47.9 & 50.4 & \underline{55.8} & \textbf{56.9} & \underline{17.3} & 11.6 & \textbf{17.6} & 15.1 \\
        \hline
        \rowcolor[HTML]{FFF2CC} Avg NLI & \underline{58.4} & 55.7 & 55.0 & \textbf{61.0} & \textbf{44.4} & 18.5 & \underline{38.6} & 36.6 \\
        \rowcolor[HTML]{F4CCCC} Avg FC & 55.6 & \underline{58.4} & 55.2 & \textbf{64.2} & \textbf{28.9} & 17.7 & 26.2 & \underline{27.6} \\
        \rowcolor[HTML]{C9DAF8} Avg HDAS & 52.8 & 57.1 & \underline{58.3} & \textbf{66.8} & 19.6 & \textbf{22.4} & 17.9 & \underline{22.1} \\
        Avg All & 56.3 & \underline{56.9} & 55.7 & \textbf{63.2} & \textbf{34.3} & 19.1 & 30.3 & \underline{30.7} \\
        \bottomrule
    \end{tabular}
    \caption{Macro F1 and Acceptability Scores on each OOD Dataset on the best models (O$^{Fev}_{128,AFk}$ and T$^{Sn}_{Full}$) and the different source dataset counterpart (T$^{Fev}_{Full}$ and O$^{Sn}_{128,AFk}$). The best score is bold, and second-best is underlined.}
    \label{tab:all_ood}
\end{table*}

\section{Discussions}
\label{sec:discuss}
This section discusses possible reasons for our earlier findings, focusing on how fine-tuning data and model influence label prediction and explanation generation. We also examine the link between label prediction performance and Acceptability scores across the three OOD tasks.

\subsection{Impact of fine-tuning dataset and base model on OOD label prediction}
\label{sec:source-f1}

\paragraph{Source dataset} To understand why models fine-tuned on the e-FEVER outperform e-SNLI on average OOD label prediction, we show the F1 score per class for both ID (in-distribution) and OOD test datasets (including cross-source and nine OOD three-label datasets) in Table~\ref{tab:id_ood_f1} in Appendix~\ref{append:comple_results}, based on O$^{Sn}_{128,AFk}$ and O$^{Fev}_{128,AFk}$ models. O$^{Sn}_{128,AFk}$ (e-SNLI) model has a better ID performance (0.86) but generalizes poorly to OOD (0.54), whereas O$^{Fev}_{128,AFk}$ (e-FEVER) model has a worse ID (0.69) but better OOD performance (0.59). For both source datasets, models perform better on e-SNLI test set, indicating that e-FEVER is harder to learn. In addition, fine-tuning on e-FEVER improved performance on classes ``Neural (NEI)'' and  ``Entailment (Supports)''.

\paragraph{Base model} We observed that T5 models exhibit more stable OOD label prediction performance than OLMo. We hypothesize this is due to: (1) T5 was fine-tuned for the supervised text-to-text language modeling objective~\cite{raffel2020exploring} including NLI datasets, and FC and HDAS are relatively similar tasks. Formatting claims (or summaries) and evidence (or documents) as hypothesis-premise pairs allows T5 to perform relatively well in few-shot settings. However, T5 showed minimal improvement with additional fine-tuning data (e.g., e-SNLI). In contrast, OLMo models started with low performance but eventually surpassed T5 as fine-tuning samples increased. (2) The T5 fine-tuning prompt aligns with its original NLI training format, requiring no adaptation. Conversely, OLMo struggled with learning from a few samples due to output formatting issues, expecting JSON with specific keys for labels and explanations. 

\subsection{Impact of fine-tuning data on OOD explanation quality}
\label{sec:source-accept}

\paragraph{Source Dataset} Models fine-tuned on e-SNLI generally have higher OOD Acceptability scores given similar F1 scores. To understand the effect of fine-tuning data on OOD explanations, Table~\ref{tab:idvsood} compares the two source datasets based on input length (hypothesis, premise, and explanations), average Acceptability scores of the original data (128 shots), and Acceptability and F1 scores for ID and OOD test sets. The results, based on O$^{Sn}_{128,AFk}$ and O$^{Fev}_{128,AFk}$, show that the input length has a large impact on the ID Acceptability score, but the impact on OOD is minor (probably due to the various OOD input length). 
Despite lower OOD F1 scores, the O$^{Sn}_{128,AFk}$ (e-SNLI) model achieves similar OOD Acceptability scores to O$^{Fev}_{128,AFk}$ (e-FEVER) model. 
This could be because part of the SNLI dataset was used to train the Acceptability model. Nevertheless, Acceptability score is more impacted by models' label prediction performance, as reflected by the F1 scores.

\begin{table}[t]
\centering
\begin{adjustbox}{width=\linewidth}
\begin{tabular}{lcccc|cc}
\toprule
 Source & \makecell{Input \\ Length} & \makecell{Source \\ Accept.} & \makecell{ID \\ Accept.} & \makecell{OOD \\ Accept.} & \makecell{ID \\ F1} & \makecell{OOD \\ F1} \\ \midrule
e-SNLI  & 38  & \textbf{0.671} & \textbf{0.565} & 0.262 & \textbf{82.8}  & 54.3  \\ 
e-FEVER & 118  & 0.394 & 0.367 & \textbf{0.263} & 58.9  & \textbf{59.9}  \\\bottomrule
\end{tabular}
\end{adjustbox}
\caption{Performance comparison across the two source datasets.}
\label{tab:idvsood}
\end{table}

\begin{figure*}[!h]
    \centering
        \includegraphics[width=\textwidth]{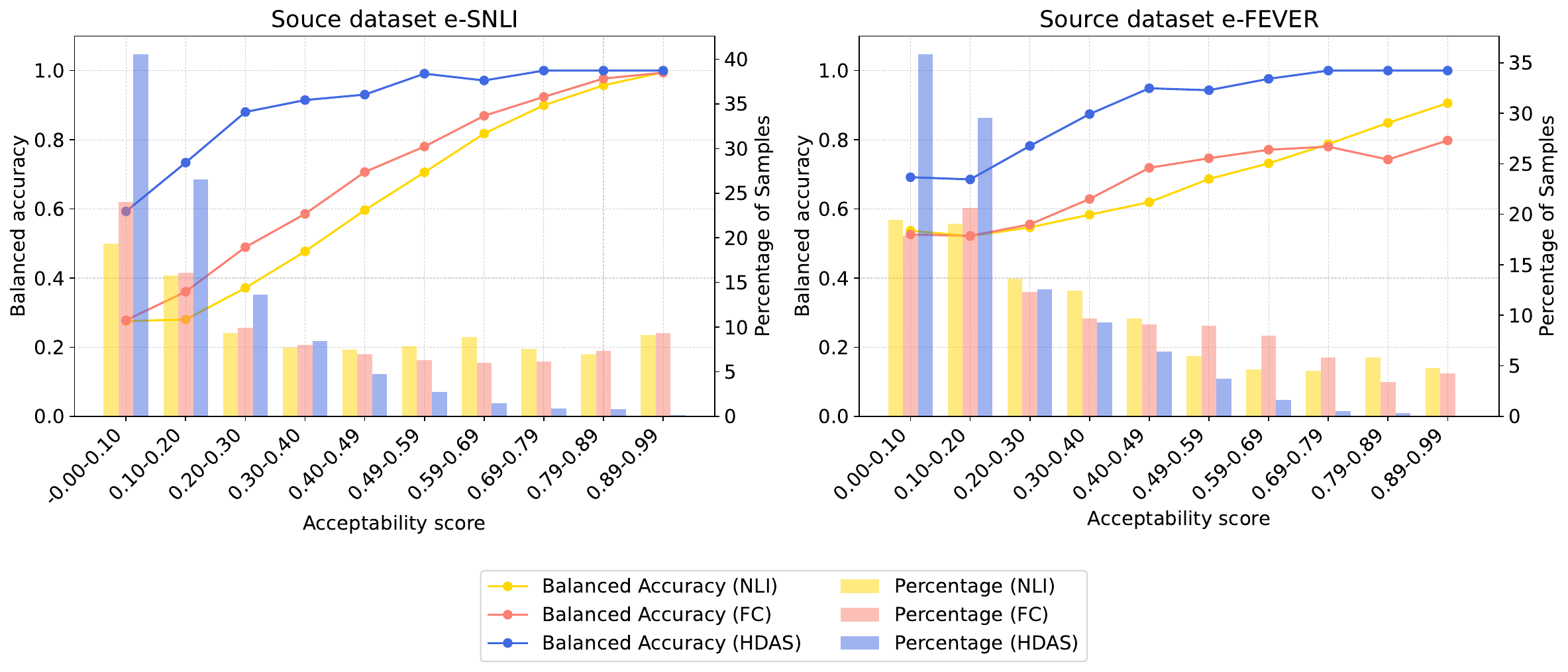}
    \caption{Distribution of label prediction accuracy (balanced) across different Acceptability score ranges. The left y-axis shows the balanced accuracy of samples from that Acceptability score range, and the right y-axis shows the percentage of samples in that range.}
    \label{fig:acc_accept}
\end{figure*}

\paragraph{Data Filtering}
Our acceptability-based (T5-Large) filtering model had only slight impacts on label prediction but improved explanation quality, according to the Acceptability score. One hypothesis is that since the Acceptability score metric (T5-11b) is a larger version of the filter model (only differing in size), the metric may favor explanations generated from models fine-tuned on acceptability-filtered samples. To investigate this, we conducted an experiment using the Themis metric as the filter for selecting samples (called "Themis-FastVote-\textit{k}"), filtering out samples with ratings below 3 (on a 1-5 scale). The experiment is based on the OLMo best model (O$^{Fev}_{128,AFk}$), and the results are shown in Table~\ref{tab:themis_filter} in Appendix~\ref{append:comple_results}. The Acceptability score with ``Themis-FastVote-\textit{k}''(0.303) is similar to ``accept-FastVote-\textit{k}''(0.307), despite having a lower F1 score. This suggests that using the acceptability filter does not cause the Acceptability metric to overestimate explanations generated from the filtered data.

\subsection{Relationship between label prediction performance and Acceptability score}

In Figure~\ref{fig:accept_f1}, we observed a positive correlation between F1 and Acceptability scores across models. We analyze on the best e-SNLI and e-FEVER models to further explore the relationship between label prediction performance and the Acceptability score within a model. We calculated the average balanced accuracy (used instead of F1 to account for varying class counts across datasets) for each task within different Acceptability score ranges, shown in Figure \ref{fig:acc_accept}. Among the three tasks, most HDAS samples have Acceptability scores below 0.3, while FC and NLI samples are distributed more evenly, indicating lower explanation quality in HDAS. When comparing source datasets, the e-SNLI model shows a steeper accuracy curve, suggesting that lower Acceptability scores often correspond to incorrect predictions of the model. In both models, the Acceptability score is positively linked to label prediction performance, especially in the lower score ranges (below 0.6).

\section{Conclusion}
This work investigated self-rationalization models' ability to generalize to NLI-related OOD tasks through the evaluation on 19 diverse datasets. We achieve this by fine-tuning T5-large and OLMo-7B under different configurations (varying fine-tuning dataset source, size, and instance selection strategies) to study the impact of data size and quality on OOD task performance and explanation quality. We also examined the Acceptability score as a reference-free metric for the generated explanation evaluation through a human evaluation. Through the study, we gained some important insights: i) fine-tuning a model on few-shot examples can perform surprisingly well in OOD datasets compared to fine-tuning on a large full-size dataset; ii) fine-tuning data source, compared to sample selection, has a larger impact on OOD performance; iii) Acceptability score is positively related to models label prediction performance. 

\section*{Limitations and Future Work}

We found that different sample selection methods had a minor impact on OOD label prediction performance, but this conclusion may not generalize to other selection methods. Our fine-tuned models were selected based on in-distribution (ID) validation sets (for T5-Large), which may limit their OOD performance, as ID and OOD performance are not always correlated. Since our OOD datasets are sourced from English-only data, this study is limited to English. Finally, with up to 128 shots, we observed performance similar to or better than full-shot models, though increasing the number of shots may yield further improvements.

Future work could explore why certain OOD datasets perform well with some models while others fail. This investigation requires a deeper understanding of distribution shifts and causality, including similarities in topics, text structure, or labeling schemes between fine-tuning and OOD~datasets. Future work could also explore ensemble learning with multiple few-shot models that may surpass full-shot fine-tuning. Another potential direction is to evaluate models' capability to generalize to multilingual OOD datasets.

\section*{Acknowledgements}
We would like to thank the anonymous reviewers and the action editor for their constructive feedback on improving this paper. We are also grateful to Haritz Puerto, Irina Bigoulaeva and Alceu Bissoto for their valuable comments and discussions.
Finally we would like to thank the funding agencies for supporting this work. Jing Yang is supported by the S\~{a}o Paulo Research Foundation (FAPESP, grant number 2019/04053-8, 2022/05002-0 and Horus project 2023/12865-8).
Max Glockner is supported by the German Federal Ministry of Education and Research and the Hessian Ministry of Higher Education, Research, Science and the Arts within their joint support of the National Research Center for Applied Cybersecurity ATHENE. This work is also supported by the LOEWE Distinguished Chair ``Ubiquitous Knowledge Processing'', LOEWE initiative, Hesse, Germany (Grant Number: LOEWE/4a//519/05/00.002(0002)/81). 

\bibliography{tacl2021}
\bibliographystyle{acl_natbib}

\appendix
\clearpage

\section{Category 1: Additional details}

\subsection{Data pre-processing} \label{append:preprocessing}
For the following datasets, we applied pre-processing as defined below:
    
\paragraph{e-FEVER} 
 We filter out incorrect explanations from e-FEVER based on the following rules (around 14\% of samples are removed from the training set):
     \begin{myitemize}
        \item The explanation is: ``The relevant information about the claim is lacking in the context.'' but the label is not NEI (NOT ENOUGH INFO).
        \item The explanation repeats the claim, and the label is not SUPPORTS. 
    \end{myitemize}

\paragraph{AddOneRTE~\cite{pavlick2016most}} We convert the mean human scores into two classes \textit{entailed} (when the score is no less than 4)  and \textit{not\_entailment} (when the score is no greater than 3, anything between 3 and 4 are removed), following the literature convention~\cite{mahabadi2020end}.

\paragraph{Ordinal Common-sense Inference (JOCI) \cite{zhang2017ordinal}} 

We follow \citet{mahabadi2020end} by mapping the labels \textit{very likely} to \textit{entailment}; \textit{likely}, \textit{plausible} and \textit{technically possible} to \textit{neutral}; and \textit{impossible} to \textit{contradiction}.

\paragraph{Multiple Premise Entailment (MPE)~\cite{lai2017natural}} We concatenate the premise sentences together to form one premise paragraph.

\paragraph{SciFact~\cite{wadden2020fact}} The dataset does not have public available labels for test set, thus we use the dev set. We do not perform evidence retrieval and use the cited document abstracts as evidence.

\paragraph{Climate FEVER~\cite{diggelmann2020climate}} 
We use the paragraph-level evidence labels.

\paragraph{FactCC~\cite{kryscinski2020evaluating}} We map label \emph{factual} as \emph{entailment} and \emph{non-factual} to \emph{not\_entailment}.

\paragraph{QAGS CNN~\cite{wang2020asking}} 
We aggregate with majority voting from the provided human annotations.

\paragraph{QAGS XSUM~\cite{wang2020asking}}
We aggregate with majority voting from the provided human annotations.

\paragraph{XSUM Hallucination~\cite{maynez2020faithfulness}} 
We aggregate with majority voting from the provided human annotations.

\subsection{Ambiguous sample selection method} \label{append:ambiguous_selection}
We input the $(h_i,p_i)$ to the T5-large model, and take the probability of the first most likely output token, since the first token represent the classification label. We denote the probability as $p_i$. 
    To select ambiguous samples, we calculate a mean probability score $p_{mean}$ as follows:
    \vspace{-2mm}
    \begin{equation}
    p_{mean} = (p_{max} + p_{min}) / 2
    \label{eq:mean}
    \end{equation}
    where $p_{max}$ and $p_{min}$ represents the highest and lowest probability score among all sample scores respectively.
    Then we re-calculate the score based on its absolute distance with $p_{mean}$:
    \begin{equation}
    p'_i = |(p_i - p_{mean})|
    \label{eq:normolize}
    \end{equation}
    with the absolute distance, we re-rank the samples from low to high to select the most ambiguous ones. The lowest value represents the most ambiguous sample and the highest the least ambiguous. 

\subsection{Additional implementation details} \label{append:implementation_details}

For T5-Large model fine-tuning, we perform a hyper-parameter search over the learning rate for each number of shots for each source dataset separately, with random sample selection from the first subset. We select the learning rate based on the highest performance on the in-distribution validation set within 50 epochs. The performance is based on the summation of label accuracy and explanation BERTscore \cite{zhang2020bertscore}. The same hyper-parameters are used for all sample selection methods, which share the same $m$ and source dataset for fine-tuning. To calculate the labels' accuracy and explanations' BERTscore, we divide the output sequence into the label and explanation. With the template format, T5 learns to generate a text label, followed by a separation pattern, ``explanation:'', and then the explanation tokens. Thus, we take the token before the separation pattern as the text label and after as the explanation. During hyper-parameter search, we test these learning rates: 3e-7, 3e-6, 3e-5, and 3e-4. 
For the validation set in fine-tuning, we randomly select 300 samples in the original validation set as the in-distribution set, as the original one is too large; thus, validation takes much longer. We follow the same settings as FEB~\cite{marasovic2022few} for the validation instances; for the ones with more than one explanation annotated, we merge them into one sequence separated by [SEP] token.

For OLMo-7B fine-tuning with LoRA, we follow recommended hyperparameters studied in~\citet{zarharan2024tell}: LoRA r and alpha values are both 16, the learning rate is 2e-4, and the optimizer is ``paged\_adamw\_32bit''.
We fine-tune all few-shot models with 50 epochs and use the models from the last epoch. For full-shot fine-tuning, the number of epochs is ten instead of 50. 

The sentence-transformer model used in embedding the input for the Fast-Vote-\textit{k} method is \textit{paraphrase-mpnet-base-v2}. 

In inference, for label mapping of T5 models, we focus on probabilities of tokens corresponding to our target labels: ``entailment'', ``contradiction'', ``neutral'', disregarding others (except for ``entailment'', as this word contains three-word tokens: ``en'', ``tail'' and ``ment'', we take the token number of ``en''). 
The label is then determined based on the highest probability among these three tokens.

\subsection{Human evaluation interface} \label{append:human_evaluation}
\begin{figure*}[h!]
        \centering
            \begin{subfigure}{0.95\textwidth}
                \centering
                \includegraphics[width=\textwidth]{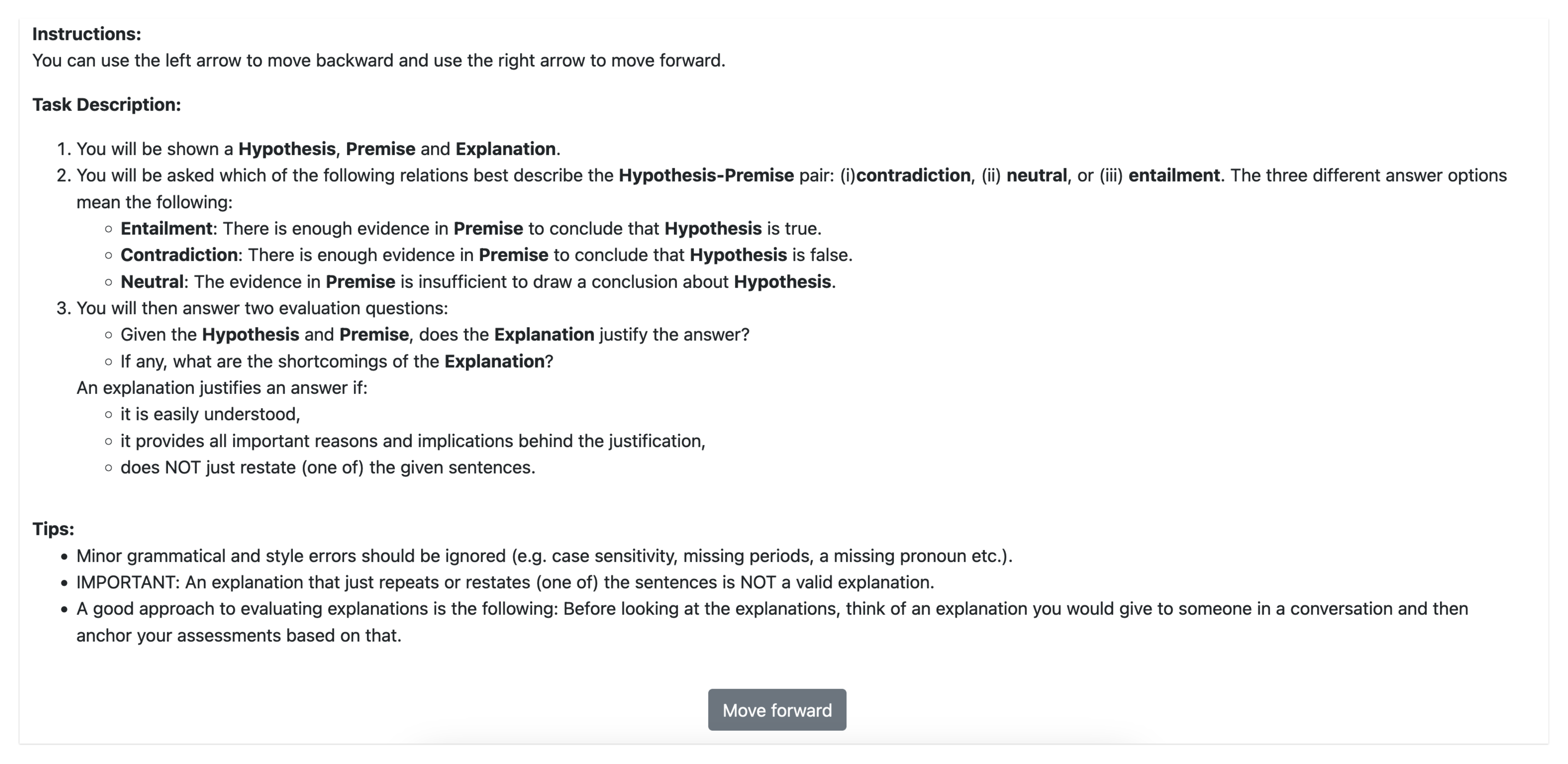}
                \caption{Task instructions}
            \end{subfigure}
            \begin{subfigure}{0.95\textwidth}
                \centering
                \includegraphics[width=\textwidth]{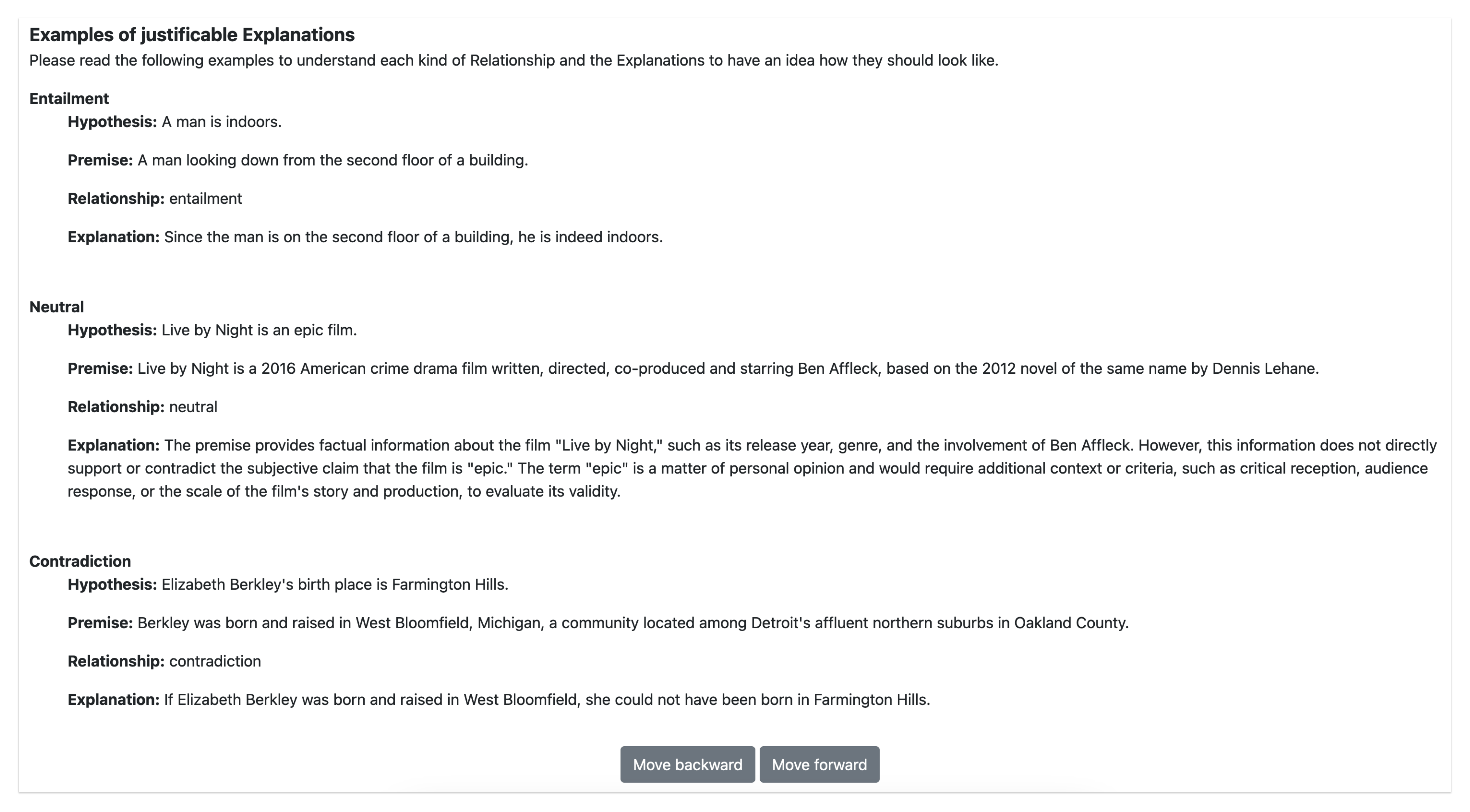}
                \caption{Examples}
            \end{subfigure}
            \begin{subfigure}{0.95\textwidth}
                \centering
                \includegraphics[width=\textwidth]{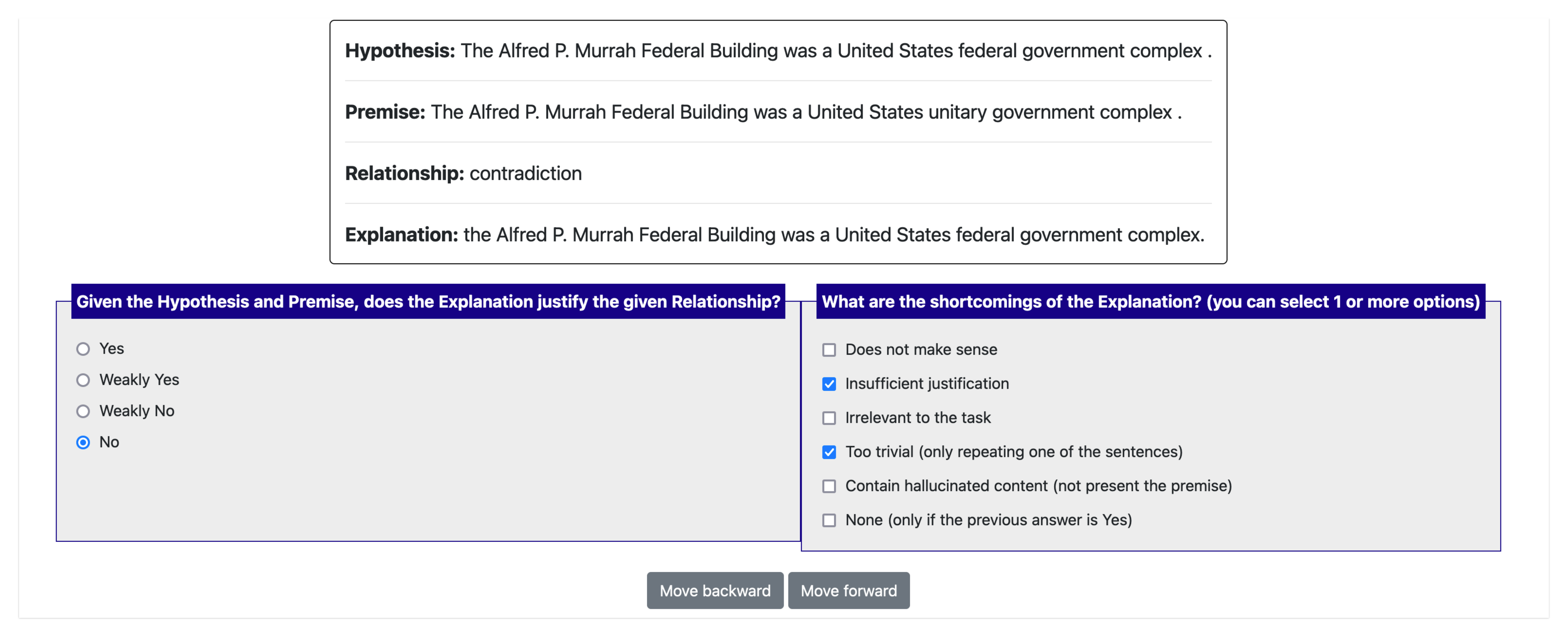}
                \caption{The evaluation page}
            \end{subfigure}
            \caption{Screenshots of human evaluation interface}
            \label{fig:example_eval}
        \end{figure*}

The evaluation interface is shown in Figure \ref{fig:example_eval}, including the task instruction, some examples, and the evaluation page. 
To select eligible participants, our screening requires participants to have at least an undergraduate degree, and primary language as English, with an approval rate above 99\%.
For high-quality evaluation, we inserted 2 attentions questions to filter out low-quality evaluations (an evaluation is rejected if the worker failed on both attention checks, or failed on one and contains invalid answers through our manual checking).

\subsection{Input template for explanation evaluation with the reference-free metrics}
\label{append:eval_template}
\begin{myitemize}
    \item \textbf{Acceptability score}
    \begin{tcolorbox}[colback=gray!5!white,colframe=white!75!black,
                  left=1mm, 
                  right=1mm] 
    \small
    \linespread{1.0}\selectfont
    \textit{premise: } [premise] ~\textit{hypothesis: } [hypothesis] ~\textit{answer: } [gold label] ~\textit{explanation: } [explanation]
    \end{tcolorbox}
    
    \item \textbf{TigerScore} and \textbf{Auto-J}
    \begin{tcolorbox}[colback=gray!5!white,colframe=white!75!black,
                  left=1mm, 
                  right=1mm] 
    \small
    \linespread{1.0}\selectfont
    \textit{Given a hypothesis and its premise, please explain why the hypothesis is entailment, neutral, or contradiction. \\
    Hypothesis:} ~[hypothesis]\textit{, Premise:} ~[premise]. \\ \textit{Please explain why the hypothesis is } ~[gold label].
    \end{tcolorbox}

    \item \textbf{Themis (relevance aspect, input in JSON format)}
    \begin{tcolorbox}[colback=gray!5!white,colframe=white!75!black,
                  left=1mm, 
                  right=1mm] 
    \small
    \linespread{1.0}\selectfont
    \{``task'': ``Controllable Generation'',
 ``aspect'': ``Coherence: Given the explanation for the relationship between the hypothesis and premise pair, how much does the generated explanation make sense?'',
 ``source\_des'': ``Hypothesis and Premise Pair'',
 ``source'': ``Hypothesis: [hypothesis], Premise: [premise], please explain why the Hypothesis is [gold label].'',
 ``target\_des'`: ``Explanation'',
 ``target'': [explanation]\}
    \end{tcolorbox}
    
\end{myitemize}

\subsection{Generated explanations by different models and their evaluation scores}
\label{append:generated_exps}

\begin{table}[ht]
\centering
\footnotesize
\begin{tabular}{@{}p{0.5\textwidth}@{}}
\toprule
\textbf{Hypothesis} (contradiction) \\ A person in a blue jacket is jumping onto a tall cement wall \\\midrule
\textbf{Premise} \\ The man is performing a large jump  \\\midrule
\textbf{T$^{Fev}_{64,AFk}$}  Human: 0.444, Accept.: 0.027                              \\
not all men wear a blue jacket.                       \\ \midrule
\textbf{T$^{Fev}_{128,R}$}  Human: 0.556, Accept.: 0.450                   \\
A man performing a large jump is not necessarily in a blue jacket and jumping onto a tall cement wall.                                \\ \midrule
\textbf{T$^{Fev}_{128,Fk}$}  Human: 0.889, Accept.: 0.036                               \\
"the relevant information about the claim is lacking in the context."                                                             \\ \midrule
\textbf{T$^{Fev}_{128,AFk}$}  Human: 1.0, Accept.: 0.494                               \\
The man is not necessarily wearing a blue jacket and the jump is not necessarily onto a tall wall. \\ \midrule
\textbf{T$^{Fev}_{Full}$ }  Human: 0.222, Accept.: 0.036                             \\
"the relevant information about the claim is lacking in the context."       \\ \midrule
\textbf{T$^{Sn}_{64,Fk}$}   Human: 0.556, Accept.: 0.520                            \\
the man is performing a large jump onto a tall cement wall.            \\ \midrule
\textbf{T$^{Sn}_{64,AFk}$}   Human: 1.0, Accept.: 0.394                            \\
the man is performing a large jump onto a tall cement wall does not indicate that he is in a blue jacket.             \\ \midrule
\textbf{T$^{Sn}_{Full}$}  Human: 1.0, Accept.: 0.842                               \\
a large jump does not imply that the man is jumping onto a tall cement wall. \\ \midrule
\textbf{O$^{Fev}_{16,AFk}$}  Human: 0.778, Accept.: 0.036                             \\
"the relevant information about the claim is lacking in the context."      \\ \midrule
\textbf{O$^{Fev}_{128,AFk}$}  Human: 0.889, Accept.: 0.473                             \\
A person in a blue jacket is performing a large jump onto a tall cement wall            \\ \midrule
\textbf{O$^{Fev}_{Full}$}   Human: 0.667, Accept.: 0.285                            \\
The man is performing a large jump, however it does not say what kind of jacket he is wearing.            \\ \midrule
\textbf{O$^{Sn}_{128,AFk}$}  Human: 0.778, Accept.: 0.536                             \\
a man in a blue jacket is performing a large jump onto a tall cement wall.            \\ \midrule
\textbf{O$^{Sn}_{Full}$}   Human: 0.667, Accept.: 0.805                            \\
The man is performing a large jump, but it does not say whether he is jumping onto a tall cement wall.         \\ \midrule
\end{tabular}%
\caption{An example of generated explanations by the 13 selected models for human evaluation, with the instance randomly selected from the SICK dataset (all models have the correct prediction).} 
\label{tab:exam_exps_sick}
\end{table}

\begin{table}[ht]
\centering
\footnotesize
\begin{tabular}{@{}p{0.5\textwidth}@{}}
\toprule
\textbf{Hypothesis} (entailment) \\ Guns N ' Roses is an American hard rock band. \\\midrule
\textbf{Premise} \\ Guns N ' Roses, often abbreviated as GnR, is an American hard rock band from Los Angeles formed in 1985.  \\\midrule
\textbf{T$^{Fev}_{64,AFk}$}  Human: 0.889, Accept.: 0.588                              \\
guns n roses is an american hard rock band.                         \\ \midrule
\textbf{T$^{Fev}_{128,R}$}  Human: 1.0, Accept.: 0.510                   \\
guns n'roses is an american hard rock band from los angeles formed in 1985.                                \\ \midrule
\textbf{T$^{Fev}_{128,Fk}$}  Human: 0.333, Accept.: 0.174                               \\
guns n'roses is an american hard rock band, however it does not say whether it is a rock band.                                                             \\ \midrule
\textbf{T$^{Fev}_{128,AFk}$}  Human: 0.778, Accept.: 0.527                               \\
guns n roses is an american hard rock band from los angeles formed in 1985. \\ \midrule
\textbf{T$^{Fev}_{Full}$ }  Human: 1.0, Accept.: 0.556                             \\
guns n'roses is an american hard rock band from los angeles formed in 1985.       \\ \midrule
\textbf{T$^{Sn}_{64,Fk}$}   Human: 0.778, Accept.: 0.498                            \\
the band was formed in 1985.            \\ \midrule
\textbf{T$^{Sn}_{64,AFk}$}   Human: 0.889, Accept.: 0.573                            \\
the band is from Los Angeles and formed in 1985.             \\ \midrule
\textbf{T$^{Sn}_{Full}$}  Human: 1.0, Accept.: 0.302                               \\
guns n'roses is an american hard rock band is a rephrasing of guns n'roses, often abbreviated as gnr, is an american hard rock band \\ \midrule
\textbf{O$^{Fev}_{16,AFk}$}  Human: 1.0, Accept.: 0.603                             \\
Guns N'Roses is an American hard rock band, formed in Los Angeles in 1985.      \\ \midrule
\textbf{O$^{Fev}_{128,AFk}$ }  Human: 0.889, Accept.: 0.556                             \\
Guns N'Roses is an American hard rock band from Los Angeles formed in 1985.            \\ \midrule
\textbf{O$^{Fev}_{Full}$ }   Human: 1.0, Accept.: 0.556                            \\
Guns N'Roses is an American hard rock band from Los Angeles formed in 1985.            \\ \midrule
\textbf{O$^{Sn}_{128,AFk}$}  Human: 0.889, Accept.: 0.618                             \\
Guns N'Roses is a hard rock band.            \\ \midrule
\textbf{O$^{Sn}_{Full}$}   Human: 0.111, Accept.: 0.088                            \\
Guns is hard to form a hard hard hard hard.         \\ \midrule
\end{tabular}%
\caption{An example of generated explanations by the 13 selected models for human evaluation, with the instance randomly selected from the VitaminC dataset (all models have the correct prediction).} 
\label{tab:exam_exps_vc}
\end{table}

\begin{table*}[ht]
\centering
\footnotesize
\begin{tabular}{@{}p{\textwidth}@{}}
\toprule
\textbf{Hypothesis} (entailment) \\ a hospital trust is being investigated by the health watchdog over its finances. \\\midrule
\textbf{Premise} \\ Monitor is looking into the financial sustainability of Southend University Hospital Foundation Trust ``on behalf of patients'', the NHS regulator said. Finances became a concern when a planned £7.8m deficit grew, for which Monitor could see no recovery plan. The hospital blamed the larger-than-anticipated deficit on growth in demand and increased staff recruitment. Hospital chairman Alan Tobias OBE, said:  ``The overspend is a result of vital investment in services and more staff to ensure high standards of patient care as well as the rise in attendances. "We welcome this review to clearly demonstrate to Monitor - as well as patients and stakeholders - our future financial plans are both robust and sustainable.'' The hospital has pledged to cut the deficit while maintaining ``high quality care for local people'' and said it was committed to returning to a surplus within three years. Monitor said its investigation will look into the state of the hospital trust\'s finances, assess the strength of its financial management and explore ways of improving its sustainability, on behalf of patients across the region. The investigation will also examine how the trust works with other local health and care organisations to respond to the financial challenges it faces. Katherine Cawley, regional director at Monitor, said: ``People are relying on Southend to provide them with quality healthcare now and in the future, and we are concerned that this may be at risk because of the trust\'s poor financial performance. "We have launched this investigation to find out more about the financial situation at the trust and to establish what can be done to improve things.''  \\\midrule
\textbf{T$^{Fev}_{64,AFk}$}  Human: 0.667, Accept.: 0.267                              \\
monitor is looking into the financial sustainability of southend university hospital foundation trust ``on behalf of patients''.                         \\ \midrule
\textbf{T$^{Fev}_{128,R}$}  Human: 0.889, Accept.: 0.276                   \\
Monitor is looking into the financial sustainability of Southend University Hospital Foundation Trust `on behalf of patients'.                               \\ \midrule
\textbf{T$^{Fev}_{128,Fk}$}  Human: 1.0, Accept.: 0.310                               \\
southend university hospital foundation trust is being investigated by monitor over its finances.                                                             \\ \midrule
\textbf{T$^{Fev}_{128,AFk}$}  Human: 0.889, Accept.: 0.319                               \\
a hospital trust is being investigated by monitor over its finances. \\ \midrule
\textbf{T$^{Fev}_{Full}$ }  Human: 0.889, Accept.: 0.266                             \\
monitor is investigating the financial sustainability of southend university hospital foundation trust ``on behalf of patients''.       \\ \midrule
\textbf{T$^{Sn}_{64,Fk}$}   Human: 0.778, Accept.: 0.240  \\ the health watchdog is looking into the financial sustainability of Southend University Hospital Foundation Trust ``on behalf of patients''
          \\ \midrule
\textbf{T$^{Sn}_{64,AFk}$}   Human: 1.0, Accept.: 0.263\\
monitor is looking into the financial sustainability of southend university hospital foundation trust ``on behalf of patients''              \\ \midrule
\textbf{T$^{Sn}_{Full}$}  Human: 0.778, Accept.: 0.219    \\
monitor is the health watchdog. \\ \midrule
\textbf{O$^{Fev}_{16,AFk}$}  Human: 1.0, Accept.: 0.262   \\
a hospital trust is being investigated by the health watchdog over its finances.      \\ \midrule
\textbf{O$^{Fev}_{128,AFk}$ }  Human: 1.0, Accept.: 0.302                             \\
The hospital trust is being investigated by the health watchdog over its finances.           \\ \midrule
\textbf{O$^{Fev}_{Full}$ }   Human: 0.889, Accept.: 0.310                            \\
southend university hospital foundation trust is being investigated by monitor over its finances.           \\ \midrule
\textbf{O$^{Sn}_{128,AFk}$}  Human: 1.0, Accept.: 0.358                             \\
Monitor is looking into the financial sustainability of Southend University Hospital Foundation Trust ``on behalf of patients'',``explanation'':``The hospital trust's poor financial performance is being investigated by the health watchdog over its finances.           \\ \midrule
\textbf{O$^{Sn}_{Full}$}   Human: 0.444, Accept.: 0.151                            \\
The financial services watch the financial policy of the financial and financial management to the financial services to the financial services.        \\ \midrule
\end{tabular}%
\caption{An example of generated explanations by the 13 selected models for human evaluation, with the instance randomly selected from the XSUM Hallucination dataset (all models have the correct prediction).} 
\label{tab:exam_exps_sxum}
\end{table*}

\clearpage

\section{Category 2: Complementary results}
\label{append:comple_results}

\begin{figure*}[htbp]
    \centering
\includegraphics[width=0.99\linewidth]{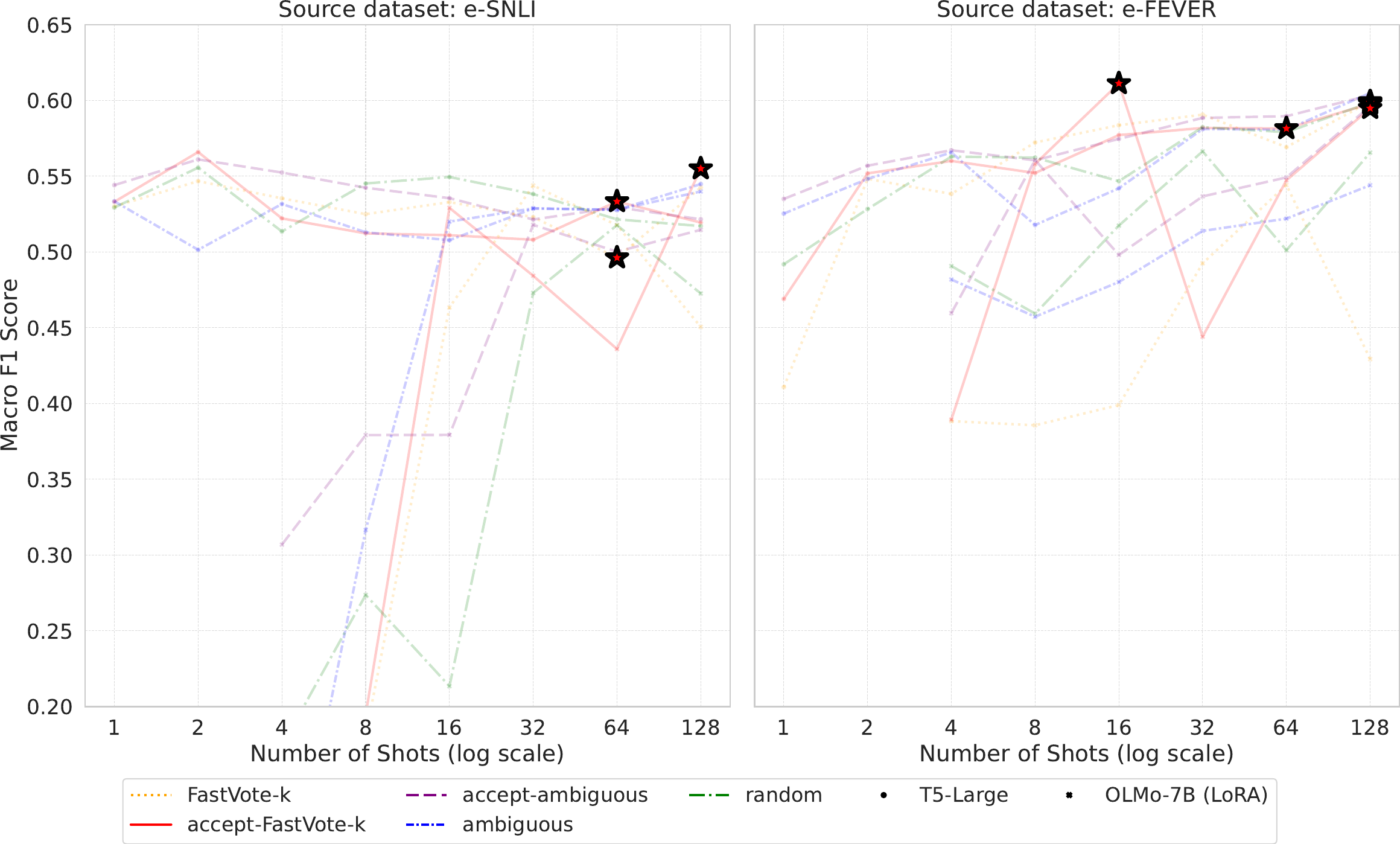}
    \caption{F1 scores of the 3 selected OOD datasets (SICK, VitaminC, XSUM Hallucination) on models fine-tuned with data from the first subset. Models marked with the asterisks are the selected ones for human evaluation (besides the full-shot models which we all include). We did not consider 1- and 2-shots fine-tuned T5 models on e-SNLI, as we observed very low quality explanations in those models.}
    \label{fig:f1_3}
\end{figure*}

\begin{figure}[H]
    \centering
    \includegraphics[width=0.5\textwidth]{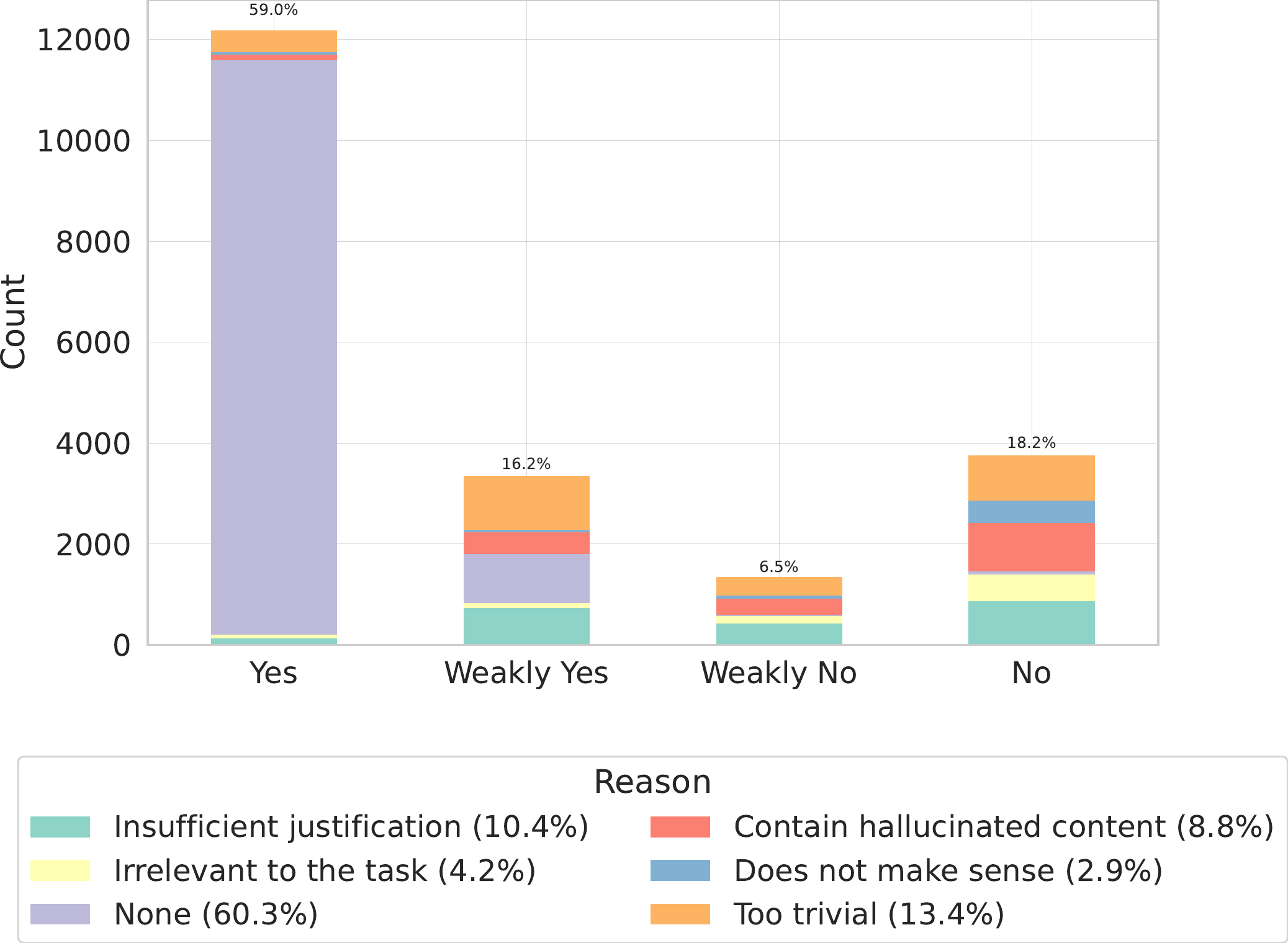}
    \caption{Distribution of reasons of shortcomings from by four answers for the question ``Does the explanation justify the answer?''. The overall explanation quality is high according to the crowd workers, around 59\% instances have ``Yes'' for the question ``Does the explanation justify the answer?''. The most common shortcoming across all answers is ``Too trivial'', followed by ``Insufficient justification'' and ``Contain hallucinated content''.}
    \label{fig:answer-reasons}
\end{figure}

\begin{table}[htbp]
        \centering
        \footnotesize
        \begin{tabular}{lrrr}
        \toprule
        Dataset & Human & Themis  & Accept.  \\
        \midrule
        SICK & \textbf{0.655} & \textbf{2.185} & \textbf{0.437} \\
        VitaminC & 0.621 & 2.183 & 0.363 \\
        XSUM H. & 0.567 & 1.633 & 0.202 \\
        \midrule
        All & 0.620 & 2.046 & 0.350 \\
        \bottomrule
        \end{tabular} 
        \caption{Human scores and automatic scores in different OOD datasets.}
        \label{tab:scores_datasets}
\end{table}

\begin{figure*}[htbp]
    \centering
\includegraphics[width=0.8\linewidth]{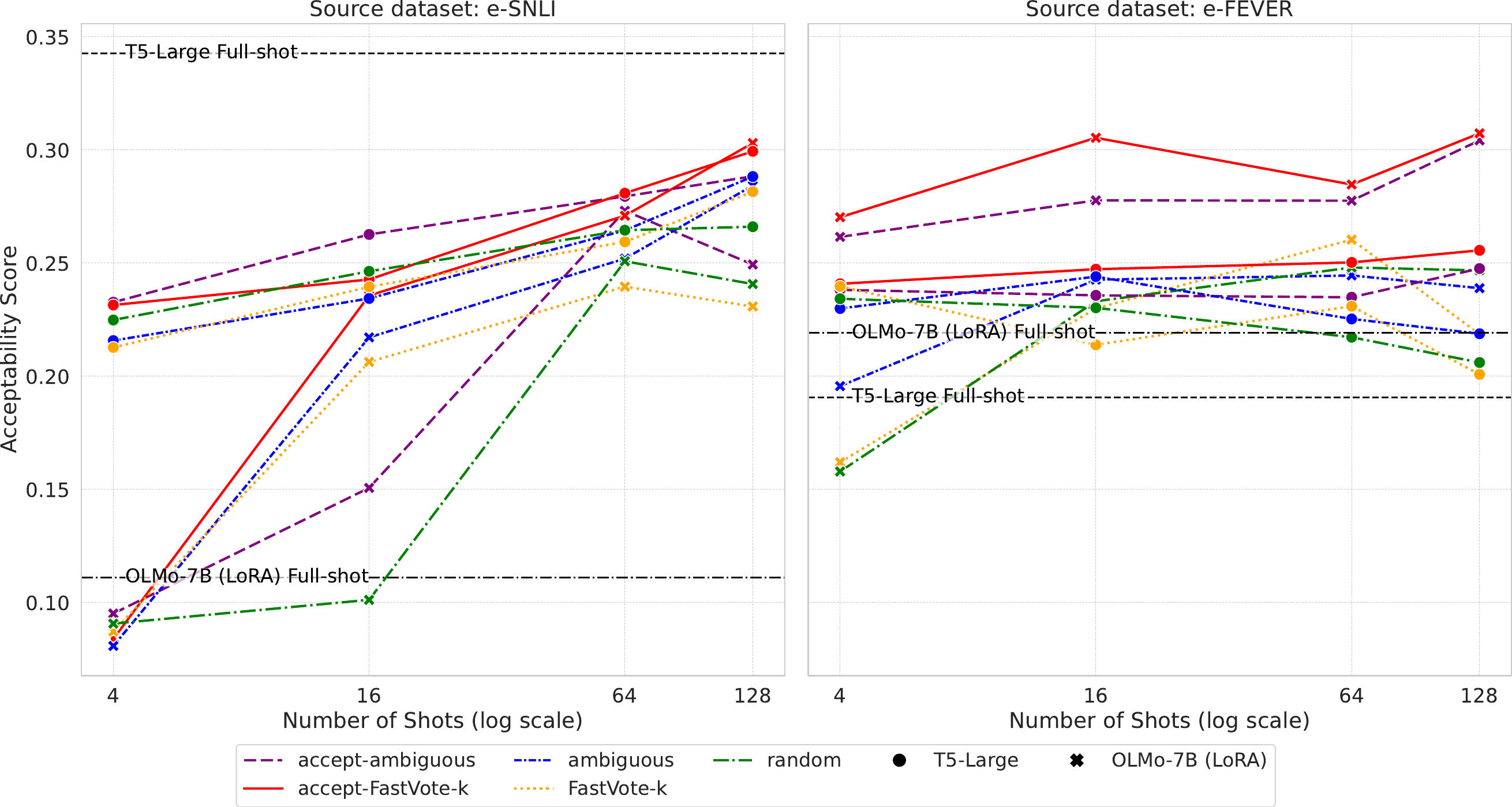}
    \caption{Acceptability score across different number of shots and sample selection~methods. Selection methods with ``accept-'' has highest Acceptability scores for all models on both source datasets.}
    \label{fig:accept_all}
\end{figure*}

\begin{table*}[htbp]
\centering
\footnotesize
\begin{tabular}{lrrrr|r|r}
\toprule
\textbf{Dataset} & \textbf{T$^{Sn}_{Full}$} & T$^{Fev}_{Full}$ & O$^{Sn}_{128,AFk}$ & \textbf{O$^{Fev}_{128,AFk}$} & \textbf{MAJ} & \textbf{SOTA} \\
\midrule
\rowcolor[HTML]{FFF2CC} SICK               & 57.1  & 82.4 & 53.7 & 64.2  & 56.9 & 90.3~\cite{chen2021neurallog} \\
\rowcolor[HTML]{FFF2CC} AddOneRTE          & 88.6  & 88.4 & 81.9 & 85.5  & 85.3 & 92.2~\cite{pavlick2016most} \\
\rowcolor[HTML]{FFF2CC} JOCI               & 53.6  & 61.5 & 47.1 & 57.9  & 57.9 & 62.6~\cite{poliak2018hypothesis} \\
\rowcolor[HTML]{FFF2CC} MPE                & 71.0  & 41.6 & 65.6 & 60.2  & 42.4 & 70.2~\cite{mahabadi2020end} \\
\rowcolor[HTML]{FFF2CC} DNC                & 60.8  & 68.3 & 55.2 & 62.1  & 50.3 & 69.0~\cite{kim2019probing} \\
\rowcolor[HTML]{FFF2CC} HANS               & 63.7  & 54.9 & 59.3 & 68.6  & 50.0 & 79.1~\cite{wu2022generating} \\
\rowcolor[HTML]{FFF2CC} WNLI               & 45.1  & 43.7 & 49.3 & 56.3  & 56.3 & 85.6~\cite{raffel2020exploring} \\
\rowcolor[HTML]{FFF2CC} Glue Diagnostics   & 60.1  & 61.9 & 58.2 & 62.7  & 41.7 & 57.0$^M$~\cite{bajaj2022metro} \\
\rowcolor[HTML]{FFF2CC} Conj               & 62.6  & 66.9 & 58.3 & 57.3  & 45.1 & 72.7~\cite{liu2023evaluating} \\
\midrule
\rowcolor[HTML]{F4CCCC} Snopes Stance      & 36.6  & 60.3 & 45.4 & 61.1  & 45.9 & 59.6$^{F1}$~\cite{hanselowski2019richly} \\
\rowcolor[HTML]{F4CCCC} SciFACT            & 65.3  & 67.7 & 54.3 & 70.0  & 41.3 & 91.4$^{F1}$~\cite{wadden2020fact} \\
\rowcolor[HTML]{F4CCCC} Climate FEVER      & 47.9  & 49.5 & 43.5 & 51.3  & 47.4 & 75.0~\cite{wolfe2024laboratory} \\
\rowcolor[HTML]{F4CCCC} VitaminC           & 59.8  & 63.0 & 58.4 & 61.0  & 50.1 & 91.1~\cite{tay2022ul2} \\
\rowcolor[HTML]{F4CCCC} COVID-Fact         & 66.5  & 74.3 & 65.1 & 76.3  & 68.3 & 83.5~\cite{saakyan2021covid} \\
\rowcolor[HTML]{F4CCCC} FM2                & 71.7  & 73.2 & 76.6 & 79.7  & 50.7 & 88.5~\cite{guan2024language} \\
\midrule
\rowcolor[HTML]{C9DAF8} FactCC             & 88.3  & 89.3 & 68.6 & 79.1  & 87.7 & 91.3$^{BA}$~\cite{yang2024reassess} \\
\rowcolor[HTML]{C9DAF8} QAGS CNN           & 75.6  & 78.2 & 62.9 & 76.8  & 74.4 & 81.3~\cite{honovich2022true} \\
\rowcolor[HTML]{C9DAF8} QAGS XSUM          & 60.3  & 62.8 & 61.5 & 72.8  & 51.5 & 77.4~\cite{honovich2022true} \\
\rowcolor[HTML]{C9DAF8} XSUM H.            & 58.9  & 62.4 & 82.9 & 80.0  & 90.1 & 66.4$^{BA}$~\cite{yang2024reassess} \\
\bottomrule
\end{tabular}
\caption{Comparison of accuracy on the 19 OOD datasets with different models. MAJ: majority voting baseline, SOTA: state-of-the-art, M: Matthews coefficient, F1: F1 score, BA: balanced accuracy.}
\label{tab:acc_comparison}
\end{table*}

\begin{table}[htbp]
\centering
\renewcommand{\arraystretch}{1.1} %
\footnotesize
\begin{tabular}{clrrrr}
\toprule
\textbf{Source} & \textbf{Test Set} & \textbf{E.} & \textbf{N.} & \textbf{C.} &\textbf{A.}\\
\midrule
\multirow{3}{*}{\rotatebox{90}{e-SNLI}} & ID (Sn)             & \textbf{86.56} & \textbf{79.62} & \textbf{91.76} & \textbf{85.98} \\
& OOD (Fev)   & 78.17 & 38.65 & 68.82 & 61.88 \\
& OOD (9)         & 59.26 & 49.56 & 51.97 & 53.60 \\
\midrule
\multirow{3}{*}{\rotatebox{90}{e-FEVER}} & ID (Fev)           & 83.22 & 48.07 & 76.39 & 69.23 \\
 & OOD (Sn)  & \textbf{89.04} & \textbf{78.18} & \textbf{86.63} & \textbf{84.61} \\
 & OOD (9)       & 69.17 & 56.64 & 52.12 & 59.31 \\
\bottomrule
\end{tabular}
\caption{F1 score performance on different test sets, contrasting the two source datasets. E.: entailment, N.: neutral, C.: contradiction, A.: average F1 score. Fev: e-FEVER, Sn: e-SNLI.}
\label{tab:id_ood_f1}
\end{table}

\begin{table}[htbp]
\centering
\footnotesize
\begin{tabular}{lccc}
\hline
\textbf{Selection}      & \textbf{Accept.} & \textbf{Themis} & \textbf{F1}  \\ \hline
Themis-FastVote-\textit{k}       & 0.303            &  3.027             & 58.24     \\ \hline
accept-FastVote-\textit{k}         & 0.307            & 2.774           & 63.24      \\ \hline
\end{tabular}
\caption{Evaluation results using Themis as a filter and as Acceptability a metric (T5-11B), compared to using acceptability as a filter (T5-Large) and Themis as a metric.}
\label{tab:themis_filter}
\end{table}

\end{document}

%% file: macros.tex
\newcommand{\mg}[1]{\textcolor{blue}{[\textbf{MG:} #1]}}
\newcommand{\jy}[1]{\textcolor{red}{[\textbf{JY:} #1]}}

\newenvironment{myenum}{
\begin{enumerate}
 \setlength{\itemsep}{1pt}
 \setlength{\parskip}{0pt}
 \setlength{\parsep}{0pt}
}{\end{enumerate}}

\newenvironment{myitemize}{
\begin{itemize}
 \setlength{\itemsep}{1pt}
 \setlength{\parskip}{0pt}
 \setlength{\parsep}{0pt}
}{\end{itemize}}